\documentclass{article} %

\usepackage{colm2024_conference}

\usepackage{booktabs}
\usepackage{graphicx}
\usepackage{enumitem}
\usepackage{wrapfig}
\usepackage{algorithm}
\usepackage{algpseudocode}
\usepackage[T1]{fontenc}

\usepackage{lmodern}

\usepackage{microtype}
\usepackage{amsmath}
\usepackage{colortbl}
\usepackage[utf8]{inputenc}
\definecolor{lightgray}{rgb}{0.9,0.9,0.9}
\definecolor{aliceblue}{rgb}{0.94, 0.97, 1.0}
\definecolor{idblue}{RGB}{30,90,150}
\usepackage{caption}
\usepackage{subcaption}
\usepackage{xcolor}
\usepackage{setspace}
\usepackage{url}
\usepackage{multirow}
\usepackage{colortbl}
\usepackage{tabularx}
\usepackage{pgfplots}
\pgfplotsset{compat=1.18} 
\usepackage{tikz}
\usetikzlibrary{er,positioning,bayesnet}
\ifdefined\pdfextension
  \newcommand{\directpdfliteral}[1]{\pdfextension literal direct {#1}}
\else
  \ifdefined\pdfliteral
    \newcommand{\directpdfliteral}[1]{\pdfliteral direct {#1}}
  \else
    \newcommand{\directpdfliteral}[1]{}
  \fi
\fi
\DeclareRobustCommand{\outlineP}{%
  \directpdfliteral{q 0 G 0.65 w [] 0 d 1 Tr}%
  P%
  \directpdfliteral{Q}%
}
\usepackage{tocloft}
\usepackage[raster,skins]{tcolorbox} %
\usepackage{xltabular}
\usepackage{adjustbox}
\usepackage{xurl}

\usepackage{amssymb}
\usepackage{mathtools}
\usepackage{amsthm}
\usepackage{enumitem}

\usepackage[table]{xcolor}

\definecolor{outlier1}{RGB}{250,252,255}
\definecolor{outlier2}{RGB}{238,244,250}
\definecolor{outlier3}{RGB}{226,236,245}
\definecolor{outlier4}{RGB}{210,225,240}
\definecolor{outlier5}{RGB}{190,210,230}
\definecolor{outlier6}{RGB}{170,195,220}
\definecolor{outlier7}{RGB}{150,180,210}

\definecolor{loss1}{RGB}{253,255,252}
\definecolor{loss2}{RGB}{251,254,250}
\definecolor{loss3}{RGB}{250,254,249}
\definecolor{loss4}{RGB}{246,252,244}
\definecolor{loss5}{RGB}{240,250,238}
\definecolor{loss6}{RGB}{234,247,232}
\definecolor{loss7}{RGB}{227,244,225}

\definecolor{gap_red1}{RGB}{255,245,205}
\definecolor{gap_red2}{RGB}{255,230,190}
\definecolor{gap_red3}{RGB}{255,210,170}
\definecolor{gap_green1}{RGB}{235,250,205}
\definecolor{gap_green2}{RGB}{220,245,190}
\definecolor{gap_green3}{RGB}{200,235,170}

\definecolor{mvio1}{RGB}{255,253,252}
\definecolor{mvio2}{RGB}{255,246,242}
\definecolor{mvio3}{RGB}{255,242,236}
\definecolor{mvio4}{RGB}{255,222,210}
\definecolor{mvio5}{RGB}{252,201,185}
\definecolor{mvio6}{RGB}{245,175,155}
\definecolor{mvio7}{RGB}{236,148,125}

\definecolor{mgray}{RGB}{232,232,236}

\author{\textbf{Peng Jin}$^*$, \textbf{Zihan Qiu}$^*$, \textbf{Zekun Wang}$^*$,  \textbf{Bo Zheng}$^*$, \\ \textbf{Yang Xu}, \textbf{Tian Xie}, \textbf{Xiao Li}, \textbf{Huaqing Zhang}, \textbf{Haoran Lian}, \textbf{Rui Men}, \\ \textbf{Dayiheng Liu}$^\dagger$
\\Qwen Team, Alibaba Token Hub, Alibaba Group \\
\small $^*$Equal contribution.\quad $^\dagger$Corresponding authors.
}
\title{ID Balancing: Stable Training of Extremely Sparse MoE via \texorpdfstring{\outlineP}{P}ID-Based Load Control}

\begin{document}

\maketitle

\begin{abstract}
Scaling Large Language Models (LLMs) via Mixture-of-Experts (MoE) enables massive parameter growth with nearly constant per-token computation. However, further scaling the parameter count requires increasingly sparse routing, where expert load imbalance becomes more severe. This imbalance reduces parameter utilization and training efficiency, and can undermine training stability, becoming a bottleneck to reliable scaling.
In this work, we unify two representative auxiliary-loss-free methods as incomplete Proportional--Integral--Derivative (PID) controllers: DeepSeek's loss-free method acts as a fixed-step integral controller, while Kimi K3's Quantile Balancing functions as a generalized proportional controller.
Building on this control perspective, we propose ID Balancing, an Integral--Derivative controller. It scales its integral term with load error and activates its derivative term only when imbalance worsens, enabling stronger corrections for large or worsening errors and smaller updates near balance. 
Evaluated across Top-$10$, Top-$5$, and Top-$3$ routing over $768$ experts, ID Balancing reduces worst-case backbone MaxVio and training-average backbone MinVio by over $50\%$ and $12\%$, respectively, relative to the best baselines in the Top-$3$ setting. When the total parameter count increases from $18.9$B to $69.9$B (Top-$10$-of-$768$), ID Balancing's worst-case backbone MaxVio remains nearly unchanged and is approximately $89.6\%$ lower than that of the auxiliary-loss baseline. ID Balancing also maintains competitive language-modeling and downstream performance. The advantages of ID Balancing grow as sparsity increases, making it a promising solution for scaling larger, sparser MoE models.

\end{abstract}

\begin{figure*}[h]
    \centering
    \includegraphics[width=\linewidth]{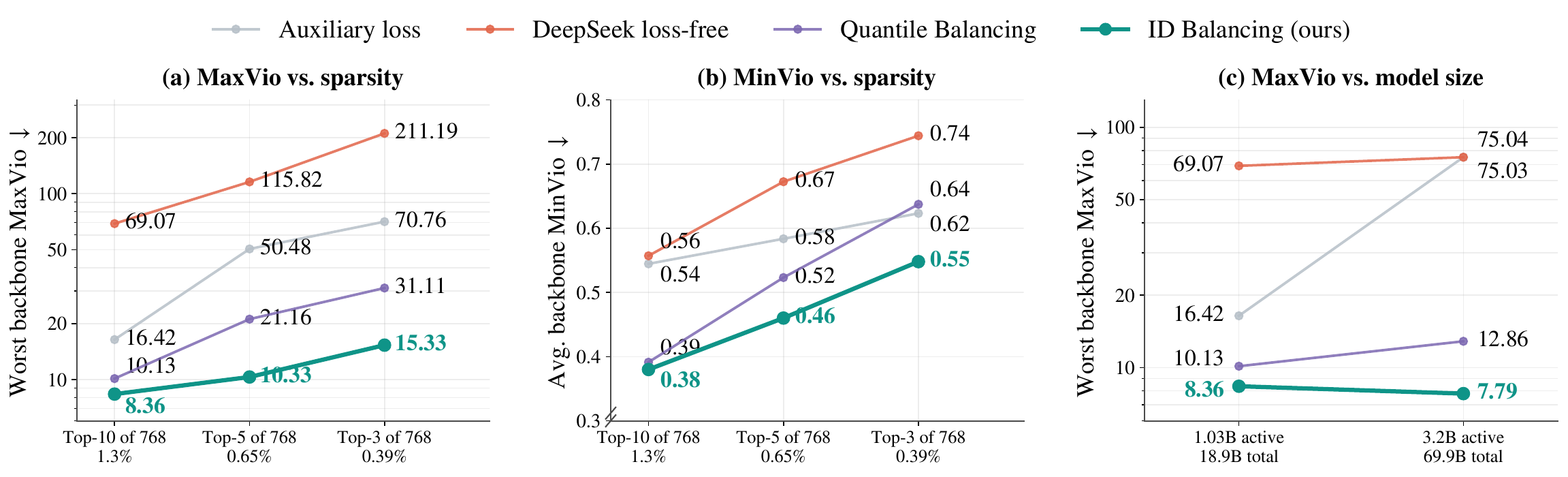}
    \caption{\textbf{ID Balancing maintains load control as routing becomes sparser and model capacity grows.} \textbf{(a)} Worst backbone MaxVio for Top-$10$, Top-$5$, and Top-$3$ routing over $768$ experts. \textbf{(b)} Training-average backbone MinVio, expressed as a decimal ratio, across the same settings. Models in (a) and (b) have $18.9$B total parameters and are trained on $120$B tokens. \textbf{(c)} Worst backbone MaxVio as total parameters increase from $18.9$B to $69.9$B ($1.03$B to $3.2$B active) under fixed Top-$10$-of-$768$ routing, using the first $30$k and $50$k steps of the small and large models, respectively.}
    \label{fig:intro-teaser}
\end{figure*}

\section{Introduction}

Mixture-of-Experts (MoE) provides an efficient way to scale large language models (LLMs) through sparse expert activation~\citep{jacobs1991adaptive,roller2021hash}. For each token, a router selects only $K$ experts from a pool of $E$. With $K$ fixed, adding experts increases total capacity without increasing the number of expert evaluations per token. This allows model capacity to grow under a fixed per-token compute budget. As the expert pool expands, the active fraction $K/E$ decreases, making extremely sparse routing a natural direction for MoE scaling~\citep{dai2024deepseekmoe,puigcerver2024sparse}.

As routing becomes sparser, maintaining balanced expert loads becomes more difficult. Small changes near the Top-$K$ selection boundary alter token assignments, making expert loads sensitive to router scores~\citep{lepikhin2020gshard,zhou2022mixture}. Overloaded experts slow down MoE computation~\citep{he2022fastermoe,he2021fastmoe,nie2022hetumoe}, while underloaded experts receive insufficient training and leave part of the model capacity unused. This imbalance limits both training efficiency and the benefits of adding more experts. Fig.~\ref{fig:intro-teaser} illustrates these challenges as routing becomes sparser or model capacity grows. Effective load control is therefore a key requirement for reliably training larger, sparser MoE models.

\begin{figure}[t]
    \centering
    \includegraphics[width=\linewidth]{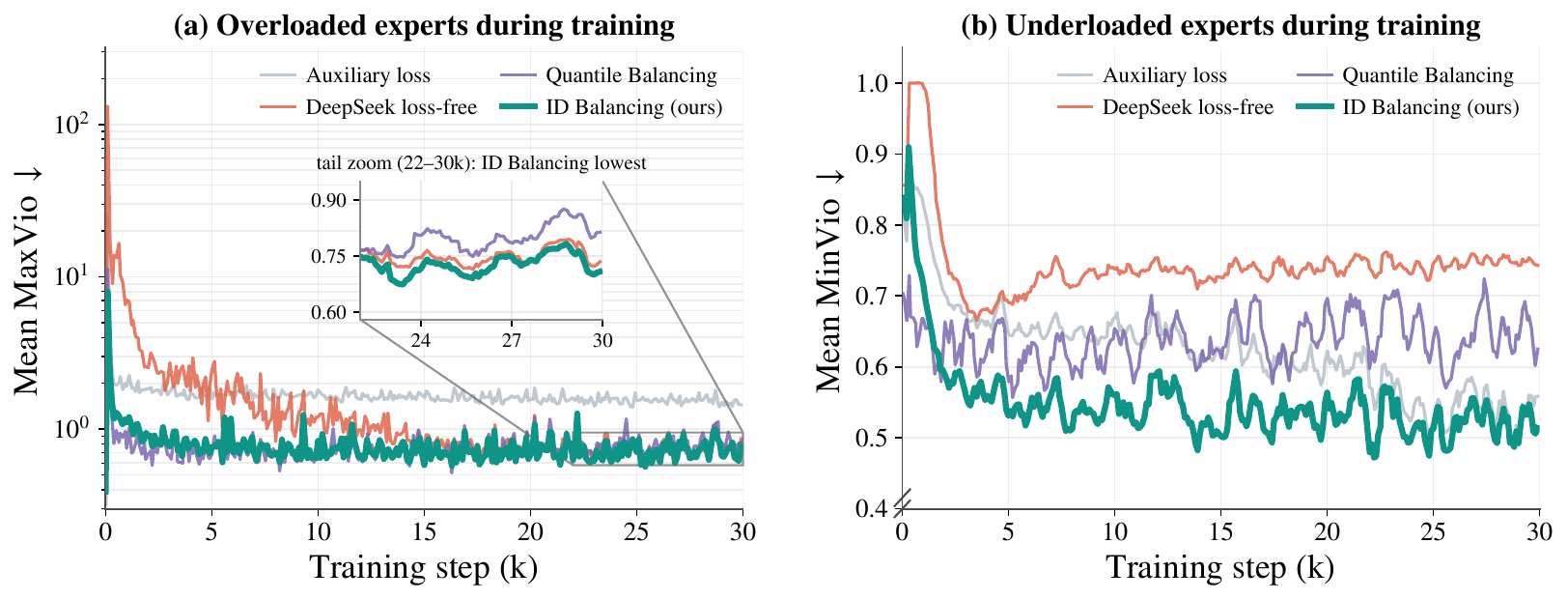}
    \caption{\textbf{ID Balancing limits transient overload and sustained underload in highly sparse MoE training.} Results use Top-$3$-of-$768$ routing in an $18.9$B-parameter model trained on $120$B tokens, with metrics averaged over $20$ backbone layers. \textbf{(a)} Mean MaxVio on a logarithmic scale ($\downarrow$). \textbf{(b)} Mean MinVio ($\downarrow$). ID Balancing achieves the lowest training-average underload.}
    \label{fig:intro_motivation}
\end{figure}

Existing auxiliary-loss-free methods~\citep{wang2024auxiliary,liu2024deepseek,team2026kimi} adjust expert biases using routing feedback. We interpret these updates through a unified Proportional--Integral--Derivative (PID) framework~\citep{johnson2005pid}, where expert biases form the control state and deviations from uniform load provide the feedback signal. DeepSeek's loss-free method acts as a fixed-step integral controller, applying the same correction to small and large load errors. Kimi K3's Quantile Balancing acts as a generalized proportional controller, setting the bias from a target computed on the current batch without explicitly accumulating past load errors. Fig.~\ref{fig:intro_motivation} illustrates the transient overload and sustained underload observed under Top-$3$-of-$768$ routing. This control perspective links their update methods to their limitations and motivates a more adaptive response: \emph{corrections should scale with the load error and respond when the imbalance worsens}.

Building on this control perspective, we propose ID Balancing, an \textbf{I}ntegral--\textbf{D}erivative controller for stable training of highly sparse MoE models. Its integral term applies larger corrections to larger load errors and smaller corrections near balance. Its derivative term uses changes in load error to add an extra correction only when the imbalance worsens. ID Balancing accumulates these corrections in the expert bias and omits the proportional term. Re-centering after each update removes the common bias component without changing routing decisions. The method requires only $\mathcal{O}(E)$ token-count feedback per layer and adds no balancing gradient to the language-modeling objective.

We evaluate ID Balancing with Top-$10$, Top-$5$, and Top-$3$ routing over $768$ experts and model sizes up to $69.9$B total parameters. We also test load control at a constant learning rate $2.3\times$ the standard peak. In the Top-$3$ setting, ID Balancing reduces worst-case backbone MaxVio~\citep{wang2024auxiliary} and training-average backbone MinVio by more than $50\%$ and $12\%$, respectively, relative to the best baselines. Under fixed Top-$10$-of-$768$ routing, its worst-case backbone MaxVio remains nearly unchanged as total parameters increase from $18.9$B to $69.9$B. These load-control improvements are accompanied by competitive language-modeling and downstream performance. Our contributions are:
\begin{itemize}[leftmargin=*]
    \item We propose a unified PID view of auxiliary-loss-free MoE balancing that connects existing update methods to their control behavior. Under this view, DeepSeek's loss-free method acts as fixed-step integral control, while Quantile Balancing acts as generalized proportional control.

    \item We propose ID Balancing, combining a magnitude-aware integral term with a worsening-gated derivative term. The method adjusts expert biases using only $\mathcal{O}(E)$ token-count feedback per layer, without adding a balancing gradient to the language-modeling objective.

    \item We validate ID Balancing across routing sparsities, model scales, and training conditions. Further analyses show effective load control during continued pretraining, improved backbone load balance, and fewer inactive experts at inference.
\end{itemize}

\section{Loss-Free Balancing as PID Control}
\label{sec:pid}

Section~\ref{sec:pid:primer} introduces the three PID branches. We then formulate MoE load balancing as a control problem in Section~\ref{sec:pid:setup} and examine the two methods in Section~\ref{sec:pid:unified}. LongCat-Flash~\citep{longcat2025technical} also adopts a PID-inspired expert-bias update, which we discuss in Appendix~\ref{app:longcat-integral}.

\subsection{The PID Controller}
\label{sec:pid:primer}

A Proportional--Integral--Derivative (PID) controller~\citep{podlubny1994fractional,ang2005pid,visioli2006practical,shah2016review,astrom2005advanced} uses feedback to drive a system toward a target. At step $t$, it computes the error $e_t$ as the target minus the measured output and produces a control signal $u_t$. The proportional branch responds to the current error, the integral branch accumulates past errors, and the derivative branch responds to changes in error~\citep{isaksson2002derivative}. A standard discrete-time form is:
\begin{equation}
    u_t = \underbrace{K_p\, e_t}_{\text{proportional}}
        + \underbrace{K_i\sum_{\tau\le t} e_\tau}_{\text{integral}}
        + \underbrace{K_d\,(e_t-e_{t-1})}_{\text{derivative}} ,
    \label{eq:pid-generic}
\end{equation}
where $K_p,K_i,K_d\ge 0$ are the gains of the three branches, and $\tau$ indexes the steps up to $t$. Eq.~\ref{eq:pid-generic} is linear in the error. For MoE balancing, we use a generalized PID view that preserves the temporal roles of the three branches while allowing nonlinear feedback transformations. This view provides a common basis for comparing how expert-bias updates use feedback over time.

\subsection{Load Balancing as a Control Problem}
\label{sec:pid:setup}

Consider an MoE layer with $E$ routed experts and a Top-$K$ router. For a token representation $\mathbf{x}\in\mathbb{R}^{d}$, the router computes logits $\mathbf{z}=W_r\mathbf{x}$, where $d$ is the hidden size and $W_r\in\mathbb{R}^{E\times d}$ is the router weight matrix. An activation function $\sigma$ maps each logit to an expert score $s_i=\sigma(z_i)$. Loss-free methods maintain a non-trainable bias $b_i$ for each expert and select experts using:
\begin{equation}
    \mathcal{T}(\mathbf{x})=\operatorname{TopK}_{i\in[E]}\!\left(s_i+b_i\right),
    \label{eq:pid-routing}
\end{equation}
where $[E]=\{1,\dots,E\}$ and $\mathcal{T}(\mathbf{x})$ contains the indices of the $K$ largest biased scores. The router obtains combination weights by normalizing the selected experts' original scores $s_i$, without the biases.

At training step $t$, let $n_i^{(t)}$ be the number of tokens assigned to expert $i$. The uniform target load and normalized load error are:
\begin{equation}
    \bar n^{(t)}=\frac{1}{E}\sum_{i=1}^{E} n_i^{(t)},
    \qquad
    e_i^{(t)}=\frac{\bar n^{(t)}-n_i^{(t)}}{\bar n^{(t)}} ,
    \label{eq:pid-error}
\end{equation}
where $\bar n^{(t)}>0$ for any nonempty batch. Negative errors indicate overload, while positive errors indicate underload. MaxVio and MinVio measure the largest deviations in these two directions, as defined in Eq.~\ref{eq:vio}. These quantities define a feedback loop with uniform load as the target. The bias vector $\mathbf{b}^{(t)}$ serves as both the persistent controller state and the routing input, while the token-count vector $\mathbf{n}^{(t)}$ is the measured output. After each step, the controller uses the error vector $\mathbf{e}^{(t)}$ to update the bias, which changes routing decisions and expert loads at the next step.

\subsection{Existing Methods through a PID Lens}
\label{sec:pid:unified}

We compare expert-bias updates by how they use routing feedback over time. Let $\mathbf{P}^{(t)},\mathbf{I}^{(t)},\mathbf{D}^{(t)}\in\mathbb{R}^{E}$ denote the proportional, integral, and derivative contributions. In a PID form, the next bias is:
\begin{equation}
    \mathbf{b}^{(t+1)}=\mathbf{P}^{(t)}+\mathbf{I}^{(t)}+\mathbf{D}^{(t)}.
    \label{eq:pid-unified}
\end{equation}
The proportional branch uses the current measurement or a target computed from the current batch. The integral branch accumulates past load errors, while the derivative branch uses the error change $\mathbf{e}^{(t)}-\mathbf{e}^{(t-1)}$. These temporal roles define the branches, allowing nonlinear feedback transformations rather than requiring the linear dependence in Eq.~\ref{eq:pid-generic}.

In MoE routing, the bias $\mathbf{b}^{(t)}$ is both a persistent controller state and the control input to the next Top-$K$ decision. Eq.~\ref{eq:pid-unified} uses the \emph{position form}, which directly gives the next bias. The same update can also be written in an \emph{incremental form} by adding $\Delta\mathbf{b}^{(t)}=\mathbf{b}^{(t+1)}-\mathbf{b}^{(t)}$ to the current bias. We use this generalized PID model as an interpretation of how a method uses feedback, not as a claim that every method follows the linear rule in Eq.~\ref{eq:pid-generic}. Under this view, DeepSeek loss-free maps to integral control, while Quantile Balancing can be interpreted as a generalized proportional controller.

\paragraph{DeepSeek's loss-free method as fixed-step integral control.}
DeepSeek's loss-free method~\citep{wang2024auxiliary,liu2024deepseek} updates each expert bias according to the sign of its load error. Starting from $\mathbf{b}^{(0)}=\mathbf{0}$, the update is:
\begin{equation}
    \mathbf{b}^{(t+1)}=\mathbf{b}^{(t)}+\eta\operatorname{sign}(\mathbf{e}^{(t)})
    =\eta\sum_{\tau\le t}\operatorname{sign}(\mathbf{e}^{(\tau)})
    =\mathbf{I}^{(t)},
    \label{eq:pid-deepseek}
\end{equation}
where $\eta>0$ is the step size and $\operatorname{sign}(\cdot)$ acts elementwise. The update raises the bias of underloaded experts and lowers that of overloaded experts. The cumulative sum in Eq.~\ref{eq:pid-deepseek} shows that the bias stores past sign errors. Under our generalized PID view, this corresponds to fixed-step integral control with $\mathbf{P}^{(t)}=\mathbf{D}^{(t)}=\mathbf{0}$. However, the sign operation discards error magnitude: every nonzero error receives a correction of size $\eta$, regardless of the severity of the imbalance.

\begin{algorithm}[t]
\caption{ID Balancing bias update, one MoE layer, per training iteration.}
\label{alg:rapid}
\begin{algorithmic}[1]
\Require expert count $E$, active experts $K$, integral gain $K_i$, derivative gain $K_d$
\State $b_i\gets 0$,\ \ $e_i^{\text{prev}}\gets 0$ \quad for all $i\in[E]$ \Comment{bias and previous error}
\For{each training iteration}
    \State route the batch by $\mathcal{T}(\mathbf{x})=\operatorname{TopK}_{i\in[E]}(s_i+b_i)$ \Comment{Eq.~\ref{eq:pid-routing}; bias frozen during the step}
    \State $n_i\gets$ number of tokens assigned to expert $i$ \Comment{summed over microbatches and the data-parallel group}
    \State $\bar n\gets\frac{1}{E}\sum_{i=1}^{E} n_i$ \Comment{uniform target load}
    \For{each expert $i\in[E]$}
        \State $e_i\gets(\bar n-n_i)/\bar n$ \Comment{normalized load error, Eq.~\ref{eq:pid-error}}
        \State $\Delta e_i\gets e_i-e_i^{\text{prev}}$
        \State $g_i\gets \mathbb{1}[\, e_i^{\text{prev}}\,\Delta e_i>0\,]$ \Comment{$1$ iff the imbalance is worsening ($e_i^{\text{prev}}$ not yet updated)}
        \State $b_i\gets b_i + K_i\, e_i + K_d\, g_i\,\Delta e_i$ \Comment{accumulate integral $+$ gated derivative}
        \State $e_i^{\text{prev}}\gets e_i$
    \EndFor
    \State $b_i\gets b_i-\frac{1}{E}\sum_{j=1}^{E} b_j$ \quad for all $i\in[E]$ \Comment{center to zero mean}
\EndFor
\end{algorithmic}
\end{algorithm}

\paragraph{Quantile Balancing as generalized proportional control.}
Quantile Balancing~\citep{team2026kimi} computes a target bias from the current batch's score distribution. Let $\mathbf{S}^{(t)}\in\mathbb{R}^{m\times E}$ contain the scores for $m$ tokens, where $S^{(t)}_{i,j}$ is the score of expert $j$ for token $i$. The method sets the next bias directly to this target:
\begin{equation}
    \mathbf{b}^{(t+1)}
    =\mathcal{Q}(\mathbf{S}^{(t)},\mathbf{b}^{(t)})
    =\mathbf{P}^{(t)}.
    \label{eq:pid-quantile}
\end{equation}
The corresponding increment, $\mathcal{Q}(\mathbf{S}^{(t)},\mathbf{b}^{(t)})-\mathbf{b}^{(t)}$, corrects the deviation from the current target.

Following Kimi K3, the target assigns approximately $q=mK/E$ tokens to each expert. For token $i$, let $\alpha_i^{(t)}$ be the $(K{+}1)$-th largest biased score $S^{(t)}_{i,r}+b_r^{(t)}$ over $r\in[E]$. The target bias for expert $j$ is:
\begin{equation}
    \mathcal{Q}_j(\mathbf{S}^{(t)},\mathbf{b}^{(t)})
    =-\operatorname{quantile}_{1-K/E}
    \big(\{\,S^{(t)}_{i,j}-\alpha_i^{(t)}\,\}_{i=1}^{m}\big),
    \label{eq:pid-quantile-def}
\end{equation}
where $S^{(t)}_{i,j}-\alpha_i^{(t)}$ is the score margin relative to the current cutoff. With these cutoffs held fixed, the target places approximately $K/E$ of the biased margins above zero, matching the desired load $q$. Ties and integer rounding affect the exact count.\footnote{At scale, we estimate the quantile from a per-expert margin histogram~\citep{team2026kimi}.} The target changes with the batch score distribution and depends on the current bias through $\alpha_i^{(t)}$. The update therefore retains state dependence, but does not explicitly accumulate past load errors. Under our generalized PID view, this direct response to a batch-dependent target forms a nonlinear proportional branch with $\mathbf{I}^{(t)}=\mathbf{D}^{(t)}=\mathbf{0}$.

\paragraph{Auxiliary loss as gradient-based balancing.}
Auxiliary-loss methods~\citep{shazeer2017outrageously,lepikhin2020gshard,xue2024openmoe} encourage balanced expert loads by adding a differentiable term to the training objective. For a batch of $T$ tokens, a common form is:
\begin{equation}
    \mathcal{L}_{\text{aux}}=\alpha\sum_{i=1}^{E} f_i\,P_i,
    \qquad
    f_i=\frac{1}{T}\sum_{\mathbf{x}}\mathbb{1}\!\left[i\in\mathcal{T}(\mathbf{x})\right],
    \qquad
    P_i=\frac{1}{T}\sum_{\mathbf{x}} p_i(\mathbf{x}),
    \label{eq:pid-auxloss}
\end{equation}
where the sums cover all tokens in the batch and $\mathbb{1}[\cdot]$ is the indicator function. Here, $f_i$ is the fraction of tokens assigned to expert $i$, and $P_i$ is its average router probability, with $p_i(\mathbf{x})=\operatorname{softmax}_i(W_r\mathbf{x})$. The coefficient $\alpha$ controls the loss strength and absorbs constant scaling factors. The balancing loss updates the router weights $W_r$ through the same optimization process as the language-modeling objective. It influences token assignments through gradients on these weights, without maintaining a separate control bias. This places auxiliary-loss balancing outside the expert-bias control family in Eq.~\ref{eq:pid-unified}.

\paragraph{Implications for load control.}
DeepSeek's loss-free method uses a fixed step regardless of error magnitude, creating a trade-off between response speed and precision. Its limited response to large errors is consistent with the early mean MaxVio spike of $132$ in Fig.~\ref{fig:intro_motivation}. Near balance, the same step risks overshooting the target. Quantile Balancing instead follows a target derived from each batch's score distribution. Changes in this target help explain its larger late-stage bias drift in Fig.~\ref{fig:method-pzero}. These observations motivate corrections that scale with the load error and respond more strongly when the imbalance worsens.

\section{ID Balancing}
\label{sec:method}

Building on the control view in Section~\ref{sec:pid:unified}, ID Balancing responds to both the magnitude and the evolution of load errors. It changes only the expert-bias update, retaining the routing rule in Eq.~\ref{eq:pid-routing} and the normalized error in Eq.~\ref{eq:pid-error}. The controller combines a magnitude-aware integral term with a derivative term that activates only when the imbalance worsens. It omits the proportional term and re-centers the bias after each update to remove its common component. Algorithm~\ref{alg:rapid} gives the complete update.

\begin{figure}[t]
    \centering
    \includegraphics[width=\linewidth]{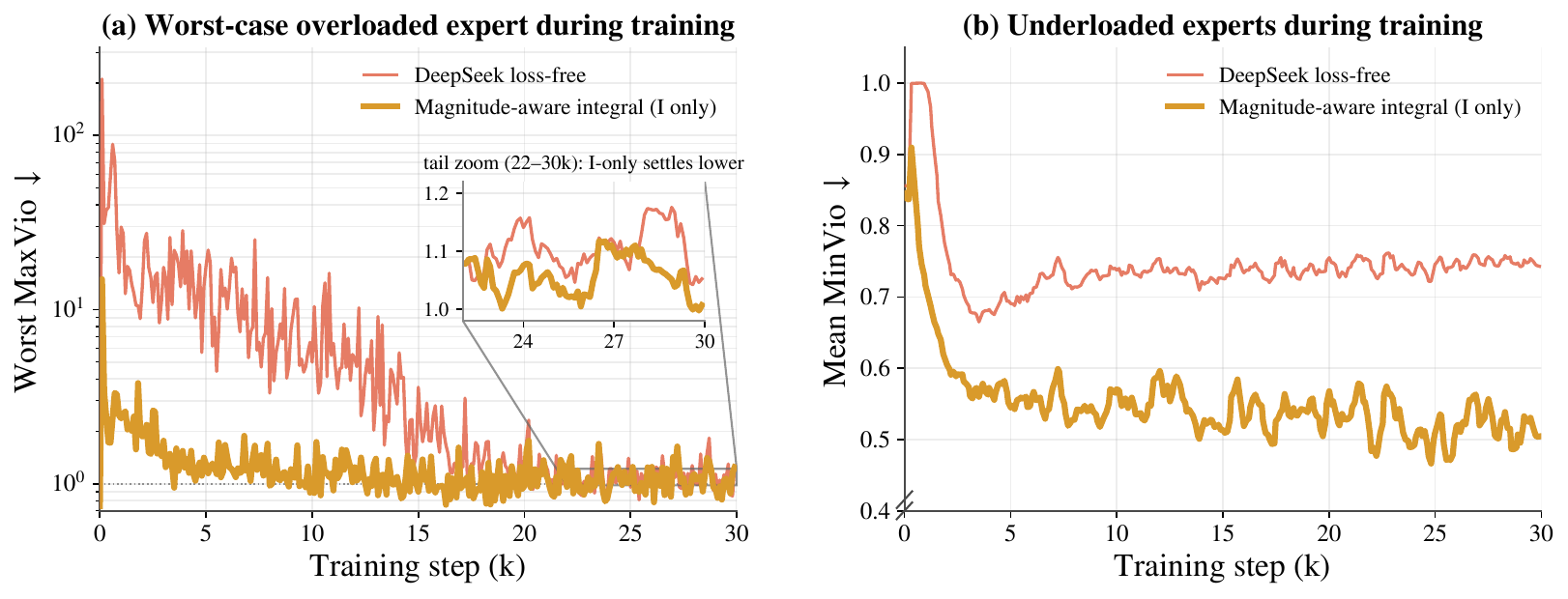}
    \caption{\textbf{Magnitude-aware integral control reduces early overload and sustained underload.} We compare the integral-only update with DeepSeek's loss-free method in an $18.9$B model using Top-$3$-of-$768$ routing and $120$B training tokens. \textbf{(a)} Maximum MaxVio across the $20$ backbone layers, shown on a logarithmic scale ($\downarrow$). \textbf{(b)} Mean MinVio across the same layers ($\downarrow$).}
    \label{fig:method-iterm}
\end{figure}

\paragraph{Magnitude-aware integral term.}
DeepSeek's loss-free method uses only the sign of the load error, giving every nonzero error the same correction size. To adapt the response to the severity of the imbalance, we scale the update by the normalized error itself:
\begin{equation}
    b_i^{(t+1)}=b_i^{(t)}+K_i\, e_i^{(t)},
    \label{eq:rapid-integral}
\end{equation}
where $K_i\ge 0$ is the integral gain shared by all experts. The correction grows with $|e_i^{(t)}|$ and shrinks as the expert approaches its target load. Since the bias accumulates these corrections, the update retains the memory of integral control. The integral update also preserves a zero-mean bias when initialized at $\mathbf{b}^{(0)}=\mathbf{0}$, because $\sum_i e_i^{(t)}=0$ by Eq.~\ref{eq:pid-error}. The fixed-sign update lacks this property: $\sum_i\operatorname{sign}(e_i^{(t)})$ is generally nonzero, allowing the common bias component to drift. Fig.~\ref{fig:method-iterm} compares the two integral updates. Scaling by error magnitude reduces the worst-layer MaxVio peak and lowers sustained underload, showing stronger early correction and better expert utilization.

\begin{figure}[t]
    \centering
    \includegraphics[width=\linewidth]{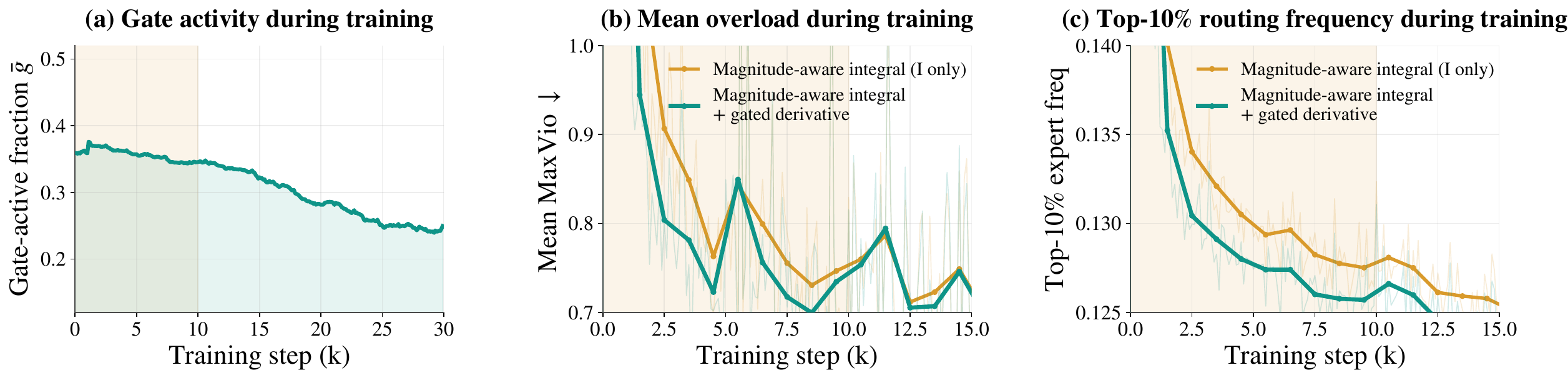}
    \caption{\textbf{The worsening-gated derivative term reduces early overload and expert concentration.} We compare magnitude-aware integral control with and without the derivative term in an $18.9$B model using Top-$3$-of-$768$ routing and $120$B training tokens. Metrics are averaged over $20$ backbone layers. \textbf{(a)} $\bar g^{(t)}$ is the fraction of active gates averaged over the 20 backbone layers. \textbf{(b)} Mean MaxVio ($\downarrow$). \textbf{(c)} Share of token assignments routed to the busiest $10\%$ of experts ($\rightarrow 0.10$), with a balanced target of $0.10$.}
    \label{fig:method-dterm}
\end{figure}

\paragraph{Worsening-gated derivative term.}
The integral term responds to error magnitude but does not distinguish growing from shrinking errors of the same size. To capture this difference, we compute the change in error:
\begin{equation}
    \Delta e_i^{(t)}=e_i^{(t)}-e_i^{(t-1)},
    \label{eq:rapid-delta}
\end{equation}
where $e_i^{(-1)}=0$. We retain only changes that move an existing imbalance farther from zero:
\begin{equation}
    g_i^{(t)}=\mathbb{1}\!\left[\, e_i^{(t-1)}\,\Delta e_i^{(t)}>0\,\right]\in\{0,1\},
    \label{eq:rapid-gate}
\end{equation}
where $\mathbb{1}[\cdot]$ is the indicator function. The gate opens when the previous error and its change have the same sign, so the imbalance grows without crossing zero. It remains closed when the error is unchanged, moves toward zero, or crosses zero. It is also closed at the first step. The correction $K_d g_i^{(t)}\Delta e_i^{(t)}$ is accumulated in the bias alongside the integral update. It uses the error difference as its input, but differs from a classical derivative output by being both gated and accumulated. Fig.~\ref{fig:method-dterm} shows that the active-gate fraction decreases from about $0.38$ early in training to $0.25$ later. Adding the gated correction lowers early load violations, while the trajectories approach those of the integral-only update as training progresses.

\begin{figure}[t]
    \centering
    \includegraphics[width=\linewidth]{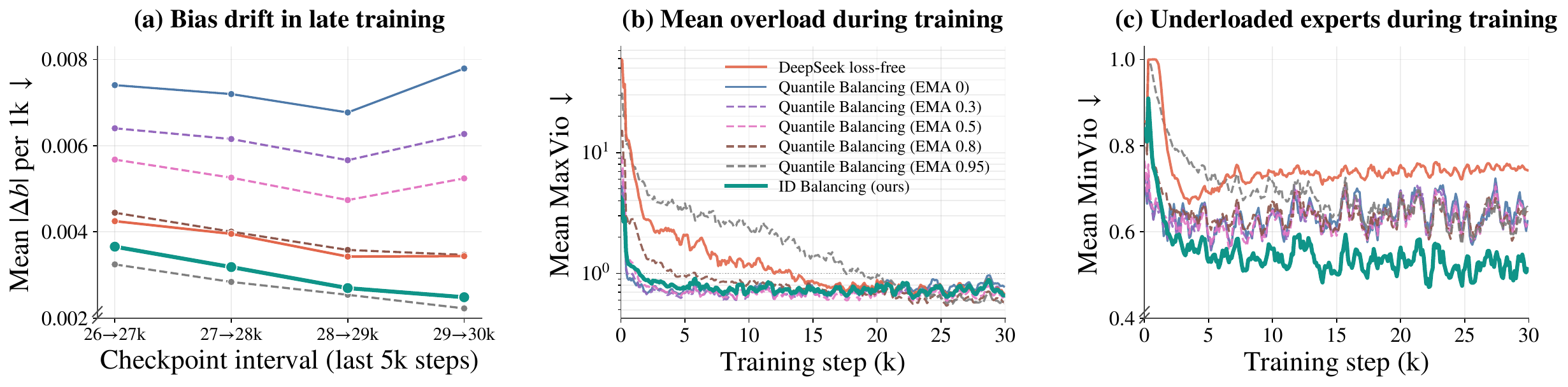}
    \caption{\textbf{EMA smoothing reduces bias drift but slows load correction in Quantile Balancing.} We compare the original update ($\rho=0$), its EMA variants, DeepSeek's loss-free method, and ID Balancing in an $18.9$B model using Top-$3$-of-$768$ routing and $120$B training tokens. All other comparisons use Quantile Balancing without EMA. \textbf{(a)} Mean absolute bias change $|\Delta b|$ between checkpoints $1$k steps apart during the final $5$k steps ($\downarrow$). \textbf{(b)} Mean MaxVio on a logarithmic scale ($\downarrow$). \textbf{(c)} Mean MinVio ($\downarrow$).}
    \label{fig:method-pzero}
\end{figure}

\paragraph{Why ID Balancing omits the proportional term.}
ID Balancing sets $K_p=0$ and controls expert loads through accumulated bias corrections. Under the generalized PID view in Section~\ref{sec:pid:unified}, Quantile Balancing follows a target computed from the current batch. As the score distribution changes, so does the target, leading to larger late-stage bias changes in our experiments. Smaller changes in relative expert biases limit perturbations to the Top-$K$ selection boundary, helping stabilize routing. Smoother bias trajectories are also desirable for weight merging, such as averaging nearby checkpoints, where the merged weights must remain compatible with the routing state. Lower bias drift may reduce this mismatch risk by keeping the routing states of the source checkpoints closer. To examine the trade-off between bias smoothing and load control, we add exponential moving average (EMA) smoothing~\citep{winters1960forecasting,hunter1986exponentially} to Quantile Balancing only as a diagnostic:
$\mathbf{b}^{(t+1)}=\rho\,\mathbf{b}^{(t)}+(1-\rho)\,\mathcal{Q}(\mathbf{S}^{(t)},\mathbf{b}^{(t)})$.
Here, $0\le\rho<1$ controls the smoothing strength, and $\rho=0$ recovers the original update without EMA.

Fig.~\ref{fig:method-pzero} shows that stronger smoothing reduces bias drift but weakens early load control. ID Balancing achieves low bias drift and effective load balance through its integral and gated derivative updates, without additional EMA smoothing.

\paragraph{Zero-mean centering step.}
The integral update has zero mean because $\sum_i e_i^{(t)}=0$. Gating does not preserve this property: $\sum_i g_i^{(t)}\Delta e_i^{(t)}$ is generally nonzero, allowing the mean bias to drift as derivative corrections accumulate. Adding the same constant to every expert bias leaves the score ranking and the selected set $\mathcal{T}(\mathbf{x})$ unchanged. We therefore remove this common component after each update. This centering removes common bias drift but does not bound the differences between expert biases.

\paragraph{The complete ID Balancing update.}
Combining the integral and gated derivative corrections with re-centering gives:
\begin{equation}
    \begin{aligned}
    \widetilde{b}_i^{(t+1)}
    &= b_i^{(t)}
    +\underbrace{K_i\,e_i^{(t)}}_{\text{magnitude-aware integral}}
    +\underbrace{K_d\,g_i^{(t)}\,\Delta e_i^{(t)}}_{\text{worsening-gated derivative}},
    \\
    b_i^{(t+1)}
    &= \widetilde{b}_i^{(t+1)}
    -\frac{1}{E}\sum_{j=1}^{E}\widetilde{b}_j^{(t+1)} .
    \end{aligned}
    \label{eq:rapid-update}
\end{equation}
where $\widetilde{b}_i^{(t+1)}$ is the bias before re-centering and $K_d\ge 0$ is the derivative gain. The second line ensures $\sum_i b_i^{(t+1)}=0$. The resulting bias controls routing at the next step.

\begin{table*}[t]
\centering
\caption{\textbf{ID Balancing improves backbone load control while maintaining competitive LM loss.} We compare Top-$3$, Top-$5$, and Top-$10$ routing over $768$ experts in $18.9$B models trained on $120$B tokens. All summaries use the same $30$k-step window. Backbone Last1k and Avg. metrics are averaged over $20$ layers, while Worst MaxVio is the maximum across layers and steps. MTP metrics are reported separately.}
\label{tab:exp-main}
\resizebox{\textwidth}{!}{%
\begin{tabular}{lc*{11}{c}}
\toprule
 & & \multicolumn{6}{c}{\textbf{Backbone}} & \multicolumn{5}{c}{\textbf{MTP Module}} \\
\cmidrule(lr){3-8}\cmidrule(lr){9-13}
\textbf{Method} & \textbf{Act./Total} & LM & \multicolumn{3}{c}{MaxVio $\downarrow$} & \multicolumn{2}{c}{MinVio $\downarrow$} & \multicolumn{3}{c}{MaxVio $\downarrow$} & \multicolumn{2}{c}{MinVio $\downarrow$} \\
\cmidrule(lr){4-6}\cmidrule(lr){7-8}\cmidrule(lr){9-11}\cmidrule(lr){12-13}
 & & loss & Last1k & Avg. & Worst & Last1k & Avg. & Last1k & Avg. & Worst & Last1k & Avg. \\
\midrule
\multicolumn{13}{l}{\cellcolor{mgray}\textbf{Top-3 of 768 experts, 120B tokens}} \\
  Auxiliary loss & \multirow{4}{*}{0.87B\,/\,18.9B} & 1.7527 & \cellcolor{mvio4}1.4809 & \cellcolor{mvio4}1.7239 & \cellcolor{mvio5}70.76 & \cellcolor{mvio3}0.5462 & \cellcolor{mvio4}0.6232 & \cellcolor{mvio5}3.8952 & \cellcolor{mvio5}4.4639 & \cellcolor{mvio5}43.38 & \cellcolor{mvio4}0.6580 & \cellcolor{mvio5}0.7502 \\
  DeepSeek loss-free &  & \cellcolor{loss7}\textbf{1.7426} & \cellcolor{mvio2}0.7314 & \cellcolor{mvio4}1.9356 & \cellcolor{mvio7}211.19 & \cellcolor{mvio5}0.7487 & \cellcolor{mvio5}0.7441 & \textbf{0.7477} & \textbf{0.8930} & \textbf{17.32} & \cellcolor{mvio4}0.6868 & \cellcolor{mvio4}0.6533 \\
  Quantile Balancing &  & \cellcolor{loss3}1.7432 & \cellcolor{mvio2}0.8093 & \cellcolor{mvio2}0.7985 & \cellcolor{mvio4}31.11 & \cellcolor{mvio4}0.6504 & \cellcolor{mvio4}0.6373 & \cellcolor{mvio4}1.8777 & \cellcolor{mvio3}1.2071 & \cellcolor{mvio4}30.89 & \cellcolor{mvio5}0.7106 & \cellcolor{mvio5}0.7083 \\
  \textbf{ID Balancing} &  & \cellcolor{loss7}\textbf{1.7426} & \textbf{0.7026} & \textbf{0.7931} & \textbf{15.33} & \textbf{0.5266} & \textbf{0.5480} & \cellcolor{mvio3}1.0165 & \cellcolor{mvio3}0.9698 & \cellcolor{mvio4}33.42 & \textbf{0.5832} & \textbf{0.5782} \\
\midrule
\multicolumn{13}{l}{\cellcolor{mgray}\textbf{Top-5 of 768 experts, 120B tokens}} \\
  Auxiliary loss & \multirow{4}{*}{0.91B\,/\,18.9B} & 1.7378 & \cellcolor{mvio3}1.0296 & \cellcolor{mvio3}1.2925 & \cellcolor{mvio5}50.48 & \cellcolor{mvio2}0.4537 & \cellcolor{mvio3}0.5836 & \cellcolor{mvio5}2.8407 & \cellcolor{mvio5}3.1892 & \cellcolor{mvio5}42.60 & \cellcolor{mvio4}0.6006 & \cellcolor{mvio4}0.6851 \\
  DeepSeek loss-free &  & \cellcolor{loss7}\textbf{1.7280} & \cellcolor{mvio1}0.5821 & \cellcolor{mvio4}1.4814 & \cellcolor{mvio6}115.82 & \cellcolor{mvio4}0.6741 & \cellcolor{mvio4}0.6725 & \textbf{0.6515} & \textbf{0.7284} & \cellcolor{mvio4}15.75 & \cellcolor{mvio3}0.5517 & \cellcolor{mvio3}0.5376 \\
  Quantile Balancing &  & \cellcolor{loss3}1.7303 & \cellcolor{mvio2}0.6376 & \textbf{0.6394} & \cellcolor{mvio4}21.16 & \cellcolor{mvio3}0.5335 & \cellcolor{mvio3}0.5232 & \cellcolor{mvio3}1.2480 & \cellcolor{mvio2}0.8683 & \textbf{7.73} & \cellcolor{mvio3}0.5516 & \cellcolor{mvio3}0.5698 \\
  \textbf{ID Balancing} &  & \cellcolor{loss5}1.7299 & \textbf{0.5404} & \cellcolor{mvio2}0.6498 & \textbf{10.33} & \textbf{0.4238} & \textbf{0.4601} & \cellcolor{mvio2}0.7808 & \cellcolor{mvio2}0.7710 & \cellcolor{mvio4}28.76 & \textbf{0.4754} & \textbf{0.4750} \\
\midrule
\multicolumn{13}{l}{\cellcolor{mgray}\textbf{Top-10 of 768 experts, 120B tokens}} \\
  Auxiliary loss & \multirow{4}{*}{1.03B\,/\,18.9B} & 1.7180 & \cellcolor{mvio2}0.6862 & \cellcolor{mvio3}1.0342 & \cellcolor{mvio4}16.42 & \cellcolor{mvio1}0.3631 & \cellcolor{mvio3}0.5446 & \cellcolor{mvio4}2.0145 & \cellcolor{mvio5}2.2781 & \cellcolor{mvio4}22.74 & \cellcolor{mvio2}0.4511 & \cellcolor{mvio3}0.5589 \\
  DeepSeek loss-free &  & \cellcolor{loss5}1.7121 & \cellcolor{mvio1}0.4418 & \cellcolor{mvio3}0.9715 & \cellcolor{mvio5}69.07 & \cellcolor{mvio3}0.5534 & \cellcolor{mvio3}0.5571 & \textbf{0.5828} & \cellcolor{mvio2}0.6066 & \cellcolor{mvio3}14.79 & \cellcolor{mvio2}0.4752 & \cellcolor{mvio2}0.4237 \\
  Quantile Balancing &  & \cellcolor{loss3}1.7133 & \cellcolor{mvio1}0.4776 & \textbf{0.4832} & \cellcolor{mvio3}10.13 & \cellcolor{mvio2}0.4060 & \cellcolor{mvio1}0.3914 & \cellcolor{mvio2}0.8642 & \cellcolor{mvio2}0.6878 & \textbf{5.16} & \textbf{0.4443} & \cellcolor{mvio2}0.4263 \\
  \textbf{ID Balancing} &  & \cellcolor{loss7}\textbf{1.7118} & \textbf{0.3863} & \cellcolor{mvio1}0.5110 & \textbf{8.36} & \textbf{0.3282} & \textbf{0.3800} & \cellcolor{mvio2}0.6070 & \textbf{0.5942} & \cellcolor{mvio4}21.98 & \cellcolor{mvio2}0.4938 & \textbf{0.3842} \\
\bottomrule
\end{tabular}%
}
\end{table*}

\section{Experiments}
\label{sec:exp}

Section~\ref{sec:exp:quality} evaluates load balance and model quality during standard and continued pretraining, followed by downstream evaluation. Section~\ref{sec:exp:stability} examines higher learning rates and optimization dynamics. We then analyze layer-wise load balance and inference-time expert utilization in Section~\ref{sec:exp:characteristics}. Finally, Section~\ref{sec:exp:ablation} isolates the effects of the integral and derivative gains and motivates the default setting $K_i=K_d=6\times10^{-3}$.

\subsection{Model Quality}
\label{sec:exp:quality}

\begin{figure}[t]
    \centering
    \includegraphics[width=\linewidth]{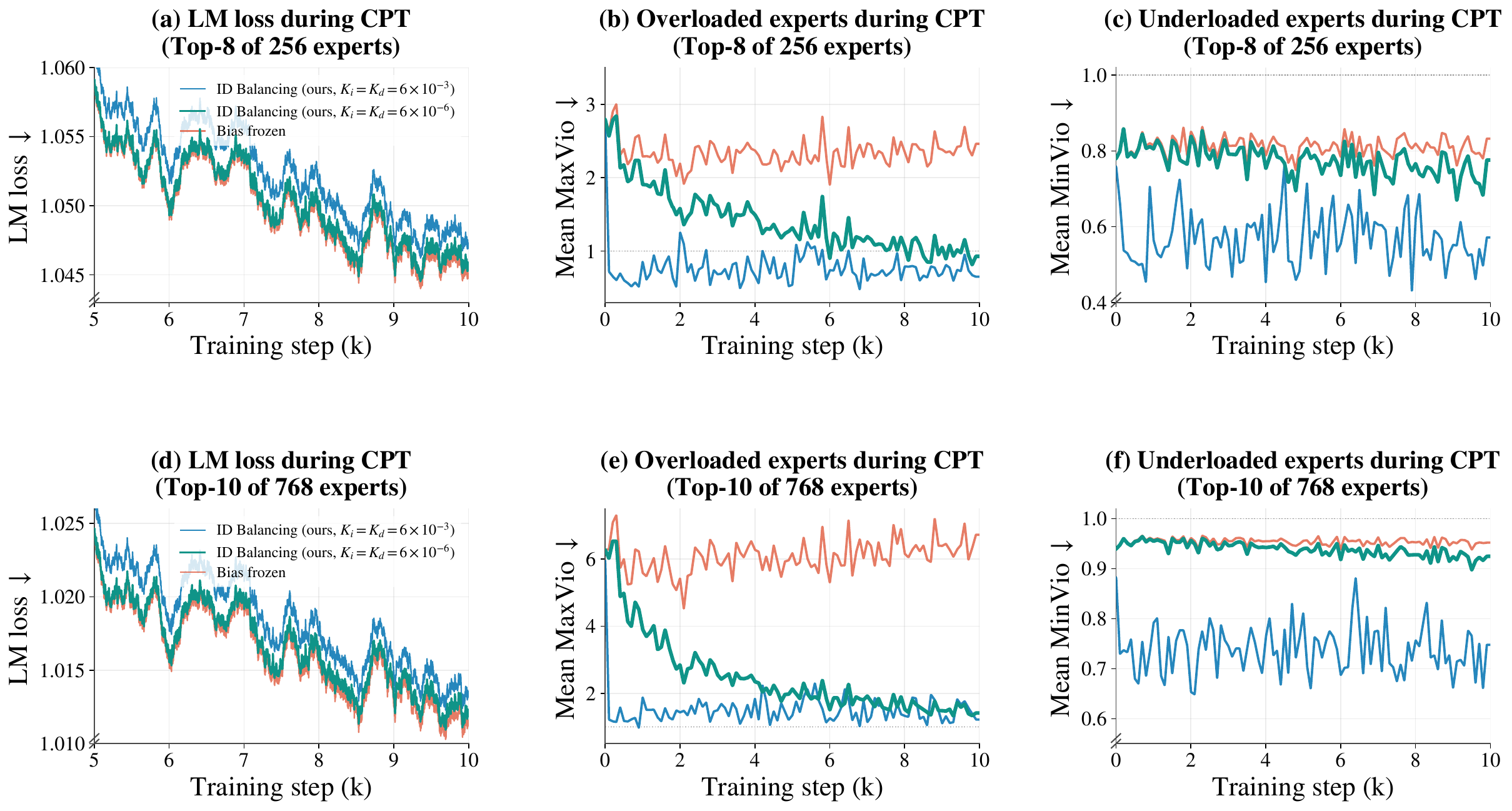}
    \caption{\textbf{Reduced ID Balancing gains limit load drift while retaining competitive LM loss during continued pretraining.} We compare full gains ($K_i=K_d=6\times10^{-3}$), gains reduced by $1000\times$ ($6\times10^{-6}$), and a frozen bias in $28$-layer models at a constant learning rate of $3\times10^{-5}$. The frozen setting matches the continued-pretraining baselines for DeepSeek's loss-free method and Quantile Balancing. The top and bottom rows use Top-$8$-of-$256$ and Top-$10$-of-$768$ routing, respectively. \textbf{(a, d)} LM loss with $201$-step smoothing ($\downarrow$). \textbf{(b, e)} Mean MaxVio ($\downarrow$). \textbf{(c, f)} Mean MinVio ($\downarrow$).}
    \label{fig:cpt-stress}
\end{figure}

\begin{figure}[t]
    \centering
    \includegraphics[width=\linewidth]{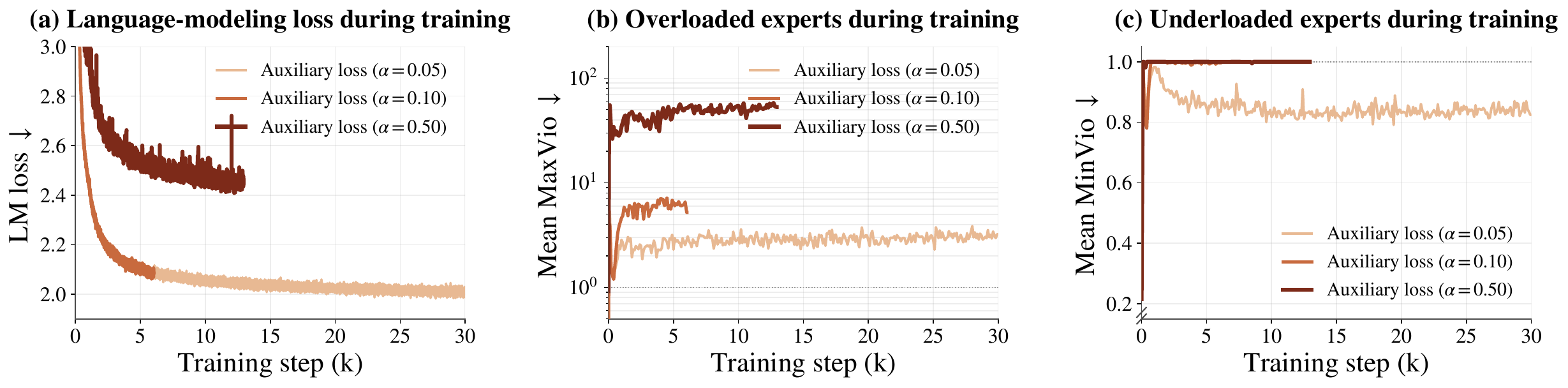}
    \caption{\textbf{Larger auxiliary-loss coefficients worsen LM loss and load balance at a higher learning rate.} We compare $\alpha\in\{0.05,0.10,0.50\}$ in a $1.03$B-active/$18.9$B-total model with Top-$10$-of-$768$ routing at a constant learning rate of $5.86\times10^{-3}$ ($2.3\times$ the standard peak). Training targets $120$B tokens, but both larger-$\alpha$ runs terminate before $30$k steps. Load metrics are averaged over $20$ backbone layers. \textbf{(a)} LM loss ($\downarrow$), with higher values at larger coefficients. \textbf{(b)} Mean MaxVio on a logarithmic scale ($\downarrow$), exceeding $50$ for $\alpha=0.50$. \textbf{(c)} Mean MinVio ($\downarrow$), reaching $1.0$ for both larger coefficients.}
    \label{fig:auxloss-stress}
\end{figure}

\paragraph{Small-model validation.}
Our main comparison uses the Qwen-3.8-Next~\citep{qiu2026design} architecture with $20$ Transformer layers, one multi-token prediction (MTP) module~\citep{gloeckle2024better}, and $768$ routed experts per MoE layer. The model has approximately $18.9$B total parameters and $0.87$--$1.03$B active parameters, depending on the routing setting. We train Top-$3$, Top-$5$, and Top-$10$ variants on $120$B tokens with a global batch size of $1024$. The learning rate peaks at $2.54\times10^{-3}$ and follows cosine decay to $3\times10^{-5}$. All comparisons use the same $30$k-step window, with backbone and MTP load metrics reported separately. We compare Auxiliary loss, DeepSeek's loss-free method, Quantile Balancing, and ID Balancing using the same architecture, data, and optimizer settings. DeepSeek's loss-free method uses a bias-update step size of $\eta=0.001$, following the DeepSeek-V3 technical report~\citep{liu2024deepseek}. The Auxiliary loss baseline uses a balancing coefficient of $\alpha=0.05$, except in the coefficient sweep in Fig.~\ref{fig:auxloss-stress}. Quantile Balancing uses its original update without EMA, except in the diagnostic experiment in Fig.~\ref{fig:method-pzero}. The main ID Balancing runs use $K_i=K_d=6\times10^{-3}$, with gain ablations in Section~\ref{sec:exp:ablation}. For continued pretraining, we reduce both gains by $1000\times$ to $6\times10^{-6}$, as examined in Fig.~\ref{fig:cpt-stress}. We measure expert overload and underload relative to the uniform target $\bar n^{(t)}$ defined in Eq.~\ref{eq:pid-error}:
\begin{equation}
    \mathrm{MaxVio}^{(t)}=\frac{\max_{i}\, n_i^{(t)} - \bar n^{(t)}}{\bar n^{(t)}} ,
    \qquad
    \mathrm{MinVio}^{(t)}=\frac{\bar n^{(t)} - \min_{i}\, n_i^{(t)}}{\bar n^{(t)}} ,
    \label{eq:vio}
\end{equation}
where $n_i^{(t)}$ is the token count of expert $i$ at step $t$. Here, $\max_i n_i^{(t)}$ and $\min_i n_i^{(t)}$ are the largest and smallest numbers of tokens assigned to any of the $E$ routed experts in the layer at step $t$, respectively. MaxVio measures the largest relative overload and is bounded by $E/K-1$. MinVio measures the largest relative deficit and lies in $[0,1]$, reaching $1$ when an expert receives no tokens. Lower values are better for both metrics. Last1k and Avg. denote means over the final $1$k steps and the full training window, respectively. Backbone values are averaged over the $20$ layers, except Worst MaxVio, which takes the maximum over both layers and steps. MTP metrics are computed separately, and LM loss is averaged over the final $1$k steps. Act./Total lists active and total parameter counts. Darker green indicates lower LM loss, darker red indicates larger load violations, and bold marks the best value in each column within a block. Tab.~\ref{tab:exp-main} shows similar LM loss among the three bias-based methods, with a maximum gap of $0.0023$, while Auxiliary loss has the highest loss in every setting. The differences in backbone load balance are larger. ID Balancing achieves the lowest Worst and Last1k MaxVio, as well as the lowest Last1k and Avg. MinVio, across all three routing settings. In Top-$3$, it reduces Worst MaxVio from $211.19$ for DeepSeek's method and $31.11$ for Quantile Balancing to $15.33$. Quantile Balancing retains lower Avg. backbone MaxVio in Top-$5$ and Top-$10$, while ID Balancing is best in Top-$3$.

\begin{table*}[t]
\centering
\caption{\textbf{ID Balancing maintains competitive downstream performance.} Results use Top-$8$-of-$256$ and Top-$10$-of-$768$ routing. Avg.\ is the mean across nine benchmarks covering knowledge, STEM, reasoning, multilingual understanding, and code generation. Block headers list active and total parameter counts. Higher scores are better, and bold marks the best result in each column within a block.}
\label{tab:exp-bench}
\setlength{\tabcolsep}{4pt}
\resizebox{\textwidth}{!}{%
\begin{tabular}{l*{10}{c}}
\toprule
 \multirow{2}{*}{\textbf{Method}} & \multicolumn{3}{c}{\textbf{Knowledge}} & \multicolumn{2}{c}{\textbf{STEM}} & \textbf{Reasoning} & \textbf{Multilingual} & \multicolumn{2}{c}{\textbf{Code}} & \multirow{2}{*}{\textbf{Avg.}} \\
\cmidrule(lr){2-4}\cmidrule(lr){5-6}\cmidrule(lr){7-7}\cmidrule(lr){8-8}\cmidrule(lr){9-10}
 & MMLU & MMLU-Pro & SuperGPQA & MATH & GSM8K & BBH & MMMLU & EvalPlus & MultiPL-E & \\
\midrule
\multicolumn{11}{l}{\cellcolor{mgray}\textbf{Top-8 of 256 experts, 3.0B\,/\,24.8B, 560B tokens}} \\
  Auxiliary loss          & 68.00 & 47.10 & 27.36 & 47.12 & 74.94 & 69.57 & 59.82 & 51.51 & 46.22 & 54.63 \\
  DeepSeek loss-free      & 70.08 & 47.23 & 27.43 & 45.84 & \textbf{76.54} & 71.09 & 60.61 & 53.96 & \textbf{46.66} & \cellcolor{loss5}55.49 \\
  Quantile Balancing      & \textbf{70.49} & 48.06 & \textbf{27.61} & \textbf{48.54} & 74.37 & 69.94 & 60.17 & 52.27 & 42.46 & \cellcolor{loss2}54.88 \\
  \textbf{ID Balancing}   & 69.58 & \textbf{48.19} & 27.52 & 47.92 & 74.49 & \textbf{72.91} & \textbf{60.78} & \textbf{56.48} & 44.40 & \cellcolor{loss7}\textbf{55.81} \\
\midrule
\multicolumn{11}{l}{\cellcolor{mgray}\textbf{Top-10 of 768 experts, 3.2B\,/\,69.9B, 560B tokens}} \\
  Auxiliary loss          & 71.67 & 50.91 & 30.15 & 50.74 & \textbf{79.08} & 71.09 & 63.62 & 56.55 & 49.28 & 58.12 \\
  DeepSeek loss-free      & 72.02 & 50.11 & 30.10 & 50.72 & 77.37 & 72.47 & 64.07 & 57.65 & \textbf{51.52} & \cellcolor{loss2}58.45 \\
  Quantile Balancing      & \textbf{72.37} & 51.14 & \textbf{30.40} & \textbf{50.86} & 75.40 & 73.42 & \textbf{64.13} & \textbf{57.94} & 51.41 & \cellcolor{loss5}58.56 \\
  \textbf{ID Balancing}   & 72.09 & \textbf{51.46} & 29.95 & 50.30 & 78.17 & \textbf{73.96} & 63.58 & 56.90 & \textbf{51.52} & \cellcolor{loss7}\textbf{58.66} \\
\bottomrule
\end{tabular}%
}
\end{table*}

\begin{table*}[t]
\centering
\caption{\textbf{ID Balancing and Quantile Balancing maintain load control at a higher learning rate.} Results use a $1.03$B-active/$18.9$B-total model with Top-$10$-of-$768$ routing, trained on $120$B tokens at a constant learning rate of $5.86\times10^{-3}$ ($2.3\times$ the standard peak). Both methods yield lower backbone load violations than Auxiliary loss and DeepSeek's loss-free method.}
\label{tab:exp-stability}
\resizebox{\textwidth}{!}{%
\begin{tabular}{l*{11}{c}}
\toprule
 & \multicolumn{6}{c}{\textbf{Backbone}} & \multicolumn{5}{c}{\textbf{MTP Module}} \\
\cmidrule(lr){2-7}\cmidrule(lr){8-12}
\textbf{Method} & LM & \multicolumn{3}{c}{MaxVio $\downarrow$} & \multicolumn{2}{c}{MinVio $\downarrow$} & \multicolumn{3}{c}{MaxVio $\downarrow$} & \multicolumn{2}{c}{MinVio $\downarrow$} \\
\cmidrule(lr){3-5}\cmidrule(lr){6-7}\cmidrule(lr){8-10}\cmidrule(lr){11-12}
 & loss & Last1k & Avg. & Worst & Last1k & Avg. & Last1k & Avg. & Worst & Last1k & Avg. \\
\midrule
  Auxiliary loss & 2.0099 & \cellcolor{mvio5}3.1383 & \cellcolor{mvio5}2.8379 & \cellcolor{mvio4}20.89 & \cellcolor{mvio6}0.8423 & \cellcolor{mvio6}0.8458 & \cellcolor{mvio5}3.5640 & \cellcolor{mvio5}3.0533 & \textbf{4.83} & \cellcolor{mvio4}0.6569 & \cellcolor{mvio5}0.7130 \\
  DeepSeek loss-free & \cellcolor{loss3}2.0031 & \cellcolor{mvio3}1.3755 & \cellcolor{mvio4}1.7811 & \cellcolor{mvio5}72.54 & \cellcolor{mvio6}0.8730 & \cellcolor{mvio6}0.8269 & \cellcolor{mvio1}0.5033 & \cellcolor{mvio2}0.7124 & \cellcolor{mvio4}17.98 & \cellcolor{mvio4}0.6105 & \cellcolor{mvio3}0.5514 \\
  Quantile Balancing & \cellcolor{loss5}2.0009 & \textbf{0.5021} & \textbf{0.4979} & \textbf{2.96} & \textbf{0.4083} & \textbf{0.3938} & \cellcolor{mvio2}0.6243 & \cellcolor{mvio1}0.5836 & \cellcolor{mvio3}10.77 & \cellcolor{mvio2}0.4346 & \cellcolor{mvio2}0.4308 \\
  \textbf{ID Balancing} & \cellcolor{loss7}\textbf{2.0007} & \cellcolor{mvio2}0.6437 & \cellcolor{mvio2}0.7274 & \cellcolor{mvio3}11.97 & \cellcolor{mvio2}0.4828 & \cellcolor{mvio2}0.4950 & \textbf{0.4642} & \textbf{0.5068} & \cellcolor{mvio3}7.20 & \textbf{0.3940} & \textbf{0.3927} \\
\bottomrule
\end{tabular}%
}
\end{table*}

\paragraph{Maintaining active load control during continued pretraining.}
DeepSeek's loss-free method freezes expert biases during continued pretraining~\citep{liu2024deepseek}. This avoids routing changes caused by bias updates but also removes active load correction~\citep{gupta2023continual}. As the model adapts to a new data distribution, expert loads may shift while the bias remains fixed~\citep{gururangan2020don,ke2023continual}. We therefore examine whether smaller ID Balancing gains preserve active load control without disrupting adaptation. We compare full gains ($K_i=K_d=6\times10^{-3}$), gains reduced by $1000\times$ ($K_i=K_d=6\times10^{-6}$), and a frozen bias in two $28$-layer models using Top-$8$-of-$256$ and Top-$10$-of-$768$ routing. Fig.~\ref{fig:cpt-stress} shows the trade-off. Full gains yield the highest LM loss, while freezing the bias leaves substantial load imbalance. With Top-$8$-of-$256$ routing, the frozen baseline reaches mean MaxVio of $2.2$--$2.5$ and MinVio of about $0.83$. With Top-$10$-of-$768$ routing, MaxVio remains between $6$ and $7$, and MinVio reaches about $0.95$, indicating severe expert underuse. The reduced gains maintain LM loss within $0.001$ of the frozen baseline during the second half of continued pretraining, while improving load balance. Mean MaxVio settles near $1$ with Top-$8$-of-$256$ routing and falls from about $6$ to $1.4$ with Top-$10$-of-$768$ routing. These results show that small bias updates correct load drift while retaining LM performance close to that of a frozen bias. We use the reduced gains as the default for continued pretraining. The DeepSeek and Quantile Balancing baselines keep their expert biases frozen during this stage.

\paragraph{Downstream quality.}
We evaluate models trained on $560$B tokens across nine benchmarks. The evaluated models use Top-$8$-of-$256$ and Top-$10$-of-$768$ routing. MMLU (5-shot)~\citep{hendrycks2020measuring}, MMLU-Pro (5-shot, Chain of Thought)~\citep{wang2024mmlu}, and SuperGPQA (5-shot, Chain of Thought) \citep{du2026supergpqa} assess general knowledge. MATH (4-shot, Chain of Thought)~\citep{hendrycks2021measuring} and GSM8K (4-shot, Chain of Thought)~\citep{cobbe2021training} assess mathematical problem solving, while BBH (3-shot, Chain of Thought)~\citep{suzgun2023challenging} evaluates reasoning. MMMLU (5-shot)~\citep{mmmlu} evaluates multilingual understanding, while EvalPlus (0-shot; the average over HumanEval~\citep{chen2021evaluating}, MBPP~\citep{austin2021program}, HumanEval+ and  MBPP+) and MultiPL-E (0-shot; Python, C++, Java, PHP, TypeScript, C\#, Bash, JavaScript)~\citep{cassano2023multipl} evaluate code generation. Tab.~\ref{tab:exp-bench} shows that ID Balancing achieves an average score of $55.81$ under Top-$8$-of-$256$ routing. Under Top-$10$-of-$768$ routing, ID Balancing improves the average score from $58.12$ to $58.66$ relative to Auxiliary loss and scores higher than this baseline on five of the nine benchmarks. These results show that ID Balancing maintains competitive downstream quality alongside improved load balance.

\begin{figure*}[t]
    \centering
    \includegraphics[width=\linewidth]{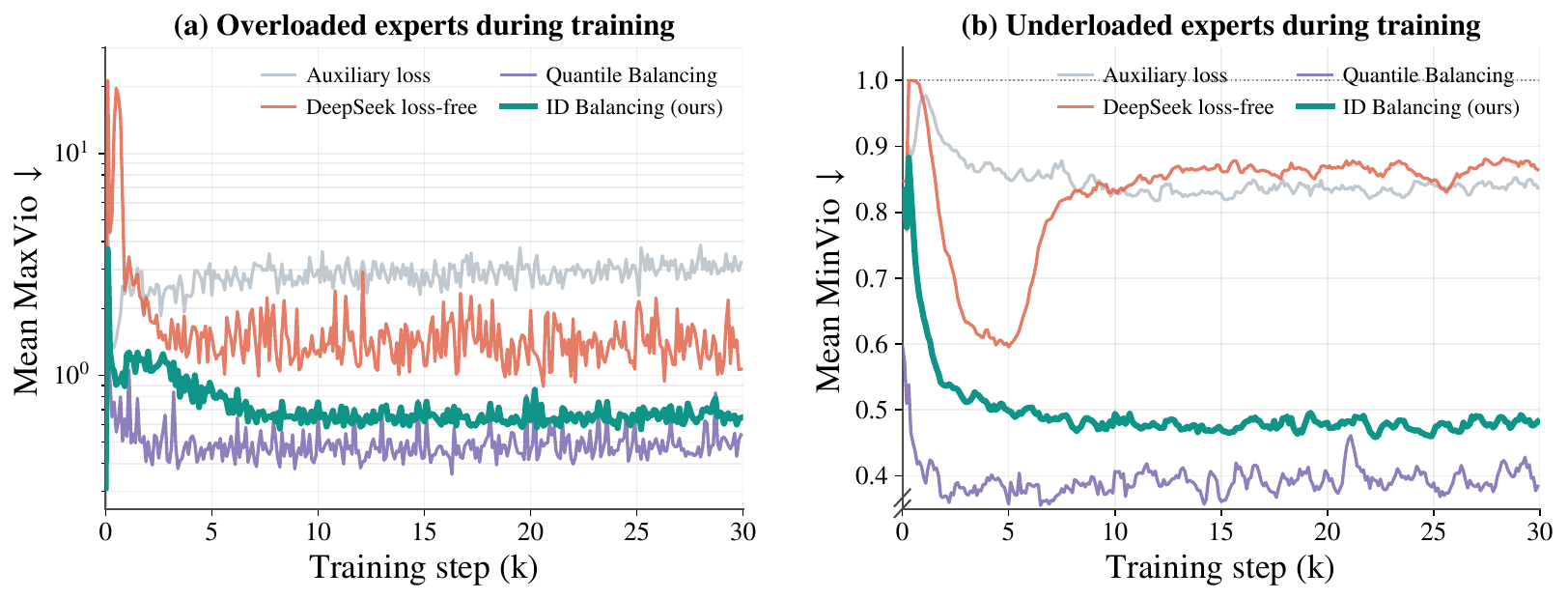}
    \caption{\textbf{ID Balancing and Quantile Balancing maintain lower backbone overload and underload at a higher learning rate.} Results use Top-$10$-of-$768$ routing and $120$B training tokens at a constant learning rate of $5.86\times10^{-3}$ ($2.3\times$ the standard peak). Metrics are averaged over $20$ backbone layers. \textbf{(a)} Mean MaxVio on a logarithmic scale ($\downarrow$), showing a large early spike for DeepSeek's loss-free method. \textbf{(b)} Mean MinVio ($\downarrow$), showing more severe underload for Auxiliary loss and DeepSeek's loss-free method.}
    \label{fig:stress-traj}
\end{figure*}

\subsection{Training Stability}
\label{sec:exp:stability}

Building on the standard-training results in Section~\ref{sec:exp:quality}, we examine load control at a higher learning rate. We also track gradient norms and MoE-output magnitudes to assess training dynamics beyond load balance.

\paragraph{Load control at a higher learning rate.}
We train the Top-$10$-of-$768$ model for $120$B tokens at a constant learning rate of $5.86\times10^{-3}$, or $2.3\times$ the standard peak. This tests load control while the learning rate remains high throughout training. Tab.~\ref{tab:exp-stability} and Fig.~\ref{fig:stress-traj} show that ID Balancing and Quantile Balancing maintain lower backbone load violations than Auxiliary loss and DeepSeek's loss-free method. Their Last1k backbone MaxVio values are $0.6437$ and $0.5021$, compared with $3.1383$ and $1.3755$ for the two baselines. Quantile Balancing gives the strongest backbone balance, while ID Balancing achieves the lowest LM loss of $2.0007$ and the lowest Last1k and Avg. MTP MaxVio of $0.4642$ and $0.5068$. These results show effective load control at the tested learning rate, with different trade-offs between backbone and MTP balance. Auxiliary loss updates the router weights $W_r$ through the same optimization process as the LM objective. Increasing its coefficient does not restore load balance in the tested range. Fig.~\ref{fig:auxloss-stress} compares $\alpha\in\{0.05,0.10,0.50\}$: larger coefficients yield higher LM loss, reaching about $2.45$ for $\alpha=0.50$, compared with about $2.01$ for $\alpha=0.05$. At $\alpha=0.50$, mean MaxVio exceeds $50$. Both larger-coefficient runs reach mean MinVio of $1.0$ and terminate before $30$k steps. The bias-based methods avoid balancing gradients, but differ in their response to load errors. DeepSeek's fixed-step update gives the same correction to every nonzero error, limiting its response to severe imbalance. Its worst backbone MaxVio reaches $72.54$. ID Balancing scales corrections with the error magnitude, consistent with the lower transient overload observed in Fig.~\ref{fig:method-iterm}. Quantile Balancing directly updates the bias toward the current batch's target and achieves the tightest backbone balance in this stress test. The comparison highlights two effective responses: error-scaled accumulation and direct batch-wise target correction.

\begin{figure*}[t]
    \centering
    \includegraphics[width=\linewidth]{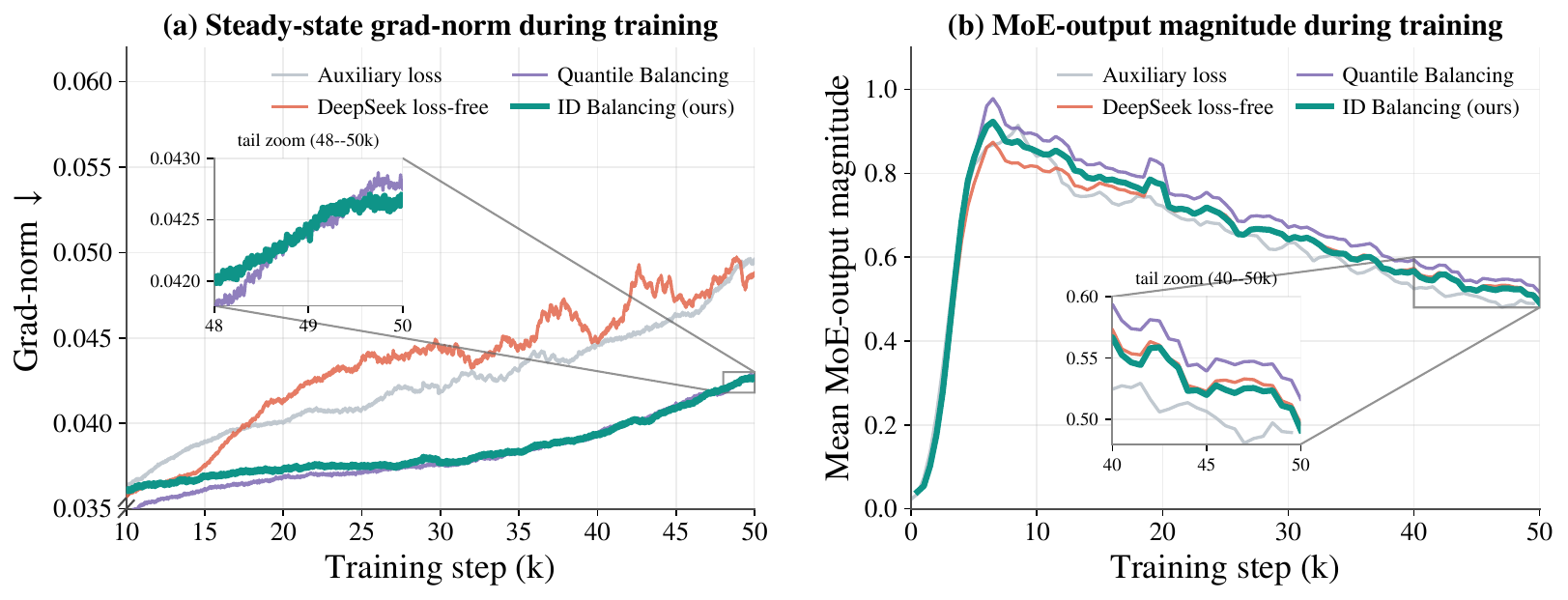}
    \caption{\textbf{Load-balancing methods show distinct gradient and activation dynamics.} Results use a $3.0$B-active/$24.8$B-total model with Top-$8$-of-$256$ routing over $50$k steps. \textbf{(a)} Global gradient norm. ID Balancing and Quantile Balancing have lower, smoother trajectories, while DeepSeek's loss-free method shows stronger late-stage fluctuations. \textbf{(b)} MoE-output magnitude averaged over the $28$ backbone layers.}
    \label{fig:gradnorm}
\end{figure*}

\paragraph{Gradient and activation dynamics.}
We examine the global gradient norm and mean MoE-output magnitude in a Top-$8$-of-$256$ model with $3.0$B active and $24.8$B total parameters. Fig.~\ref{fig:gradnorm} reports both signals, with output magnitudes averaged over the $28$ backbone layers~\citep{wang2024deepnet,xiong2020layer}. ID Balancing and Quantile Balancing end with gradient norms near $0.043$, below Auxiliary loss at $0.050$ and DeepSeek's loss-free method at $0.049$. DeepSeek's method also shows stronger late-stage fluctuations. MoE-output magnitudes rise early and then decline, with Quantile Balancing ending near $0.54$ and the other methods near $0.51$--$0.52$. ID Balancing thus maintains a lower, smoother gradient norm while retaining an output scale close to those of Auxiliary loss and DeepSeek's loss-free method.

\begin{figure*}[t]
    \centering
    \includegraphics[width=\linewidth]{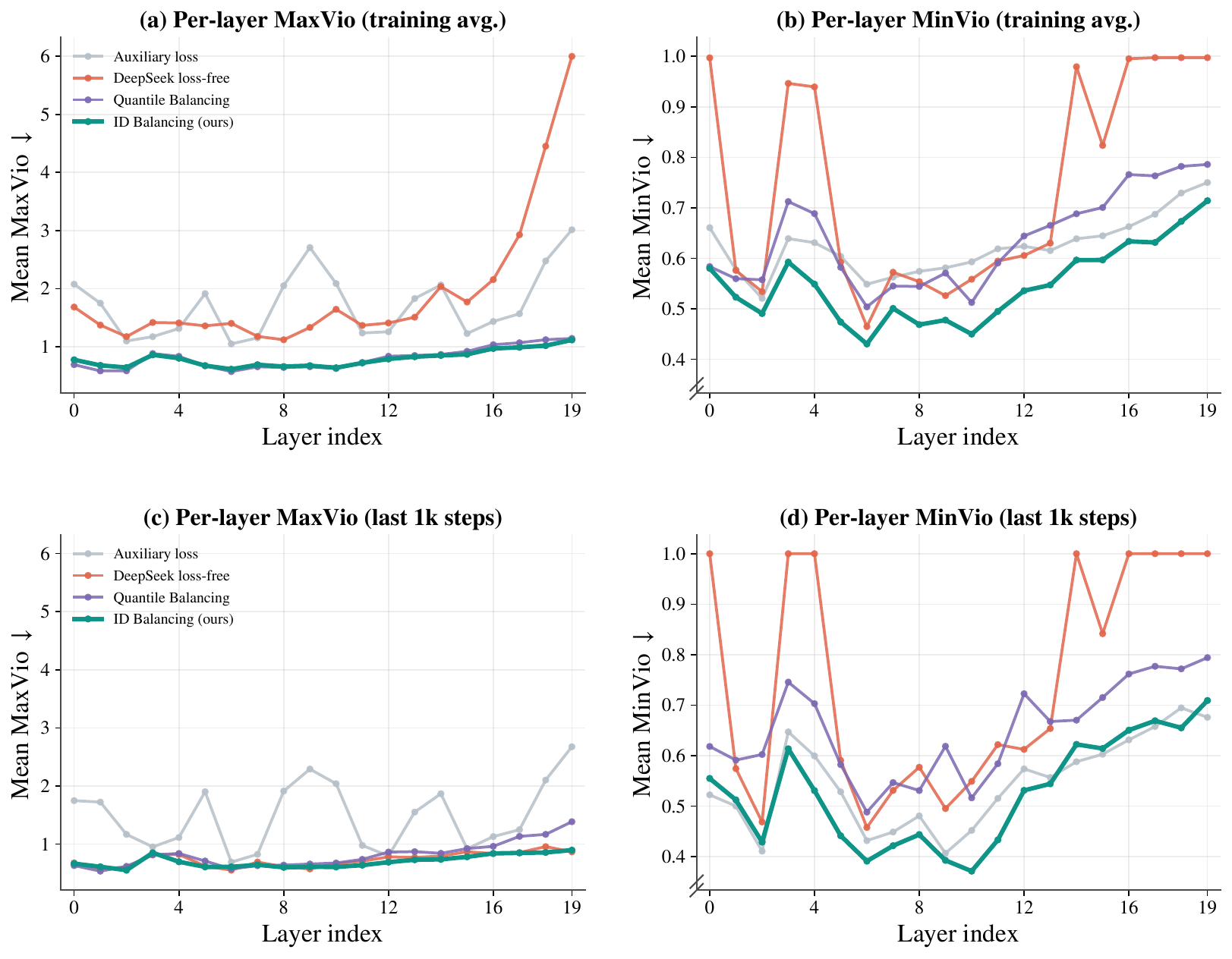}
    \caption{\textbf{ID Balancing maintains consistent overload and underload control across layers.} Results use Top-$3$-of-$768$ routing over $30$k steps and cover all $20$ backbone layers. The top and bottom rows show the training average and the final $1$k-step average, respectively. \textbf{(a, c)} MaxVio ($\downarrow$). \textbf{(b, d)} MinVio ($\downarrow$). ID Balancing achieves lower or comparable MaxVio and lower MinVio than Quantile Balancing across all layers. MinVio generally increases with depth for all four methods.}
    \label{fig:perlayer-vio}
\end{figure*}

\subsection{Model Characteristics}
\label{sec:exp:characteristics}

\paragraph{Training-time characteristics.}
Fig.~\ref{fig:perlayer-vio} complements the aggregate results in Tab.~\ref{tab:exp-main} by reporting MaxVio and MinVio for all $20$ backbone layers under Top-$3$-of-$768$ routing. The top and bottom rows show the training average and the final $1$k-step average, respectively. Auxiliary loss shows substantial variation in both metrics across layers, with only modest improvement in the final $1$k steps. DeepSeek's loss-free method has particularly high training-average overload in deeper layers: MaxVio reaches about $6.0$ at layer $19$, compared with $0.9$ over the final $1$k steps. Despite this lower late-stage overload, MinVio remains near $1.0$ in several layers, indicating persistent underutilization. Quantile Balancing and ID Balancing maintain more consistent load control across layers, with smaller differences between their training-average and final-stage profiles. ID Balancing achieves lower or comparable MaxVio and lower MinVio than Quantile Balancing across all $20$ layers. Its aggregate advantage therefore reflects improvements throughout the backbone. For all four methods, MinVio generally increases in deeper layers. ID Balancing limits underload but does not eliminate this shared depth-wise trend.

\begin{figure*}[t]
    \centering
    \includegraphics[width=\linewidth]{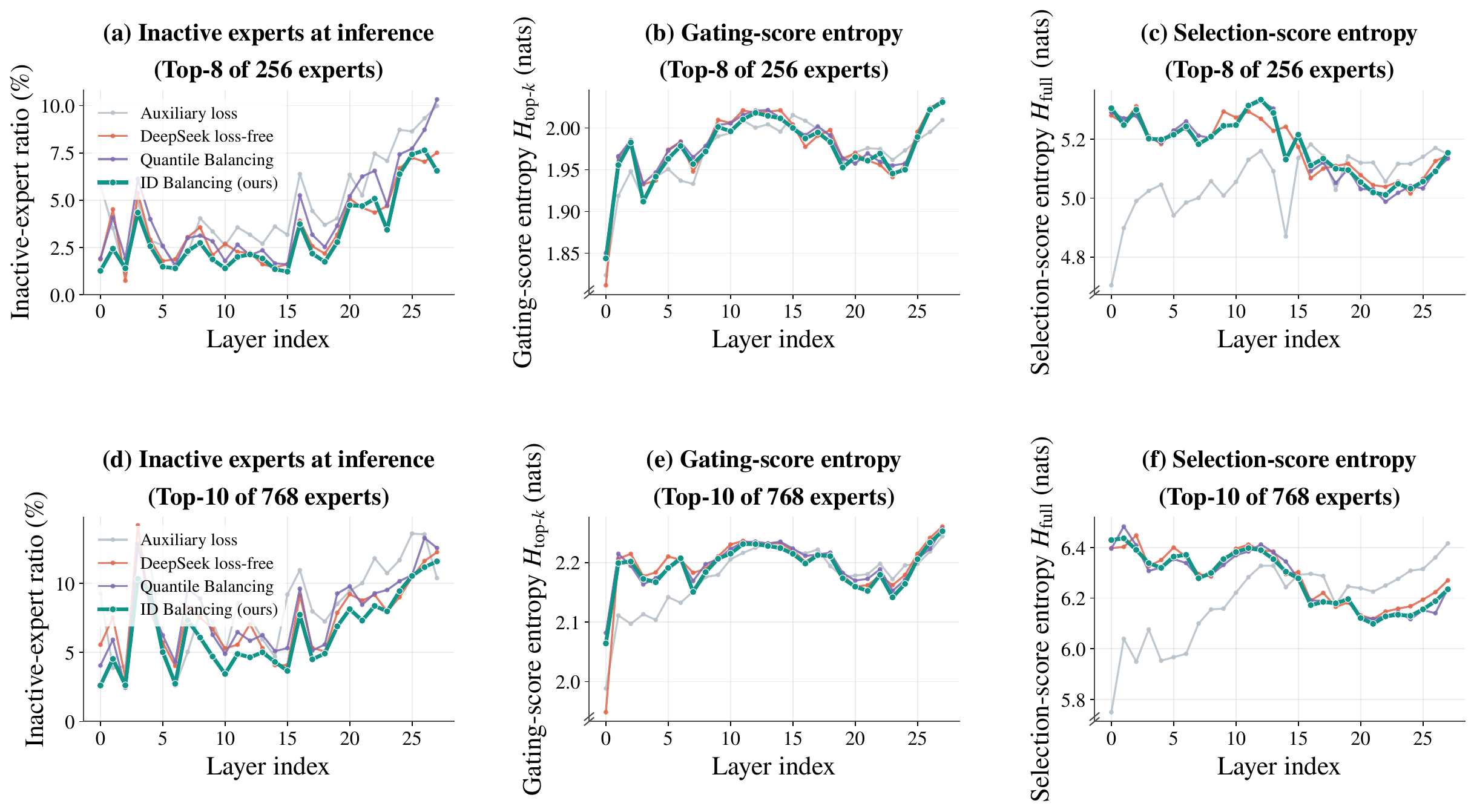}
    \caption{\textbf{ID Balancing reduces expert inactivity relative to Auxiliary loss.} Results are averaged by layer over nine downstream benchmarks. The top and bottom rows use Top-$8$-of-$256$ and Top-$10$-of-$768$ routing, respectively. \textbf{(a, d)} Inactive-expert ratio, the fraction of experts receiving no evaluation token ($\downarrow$). \textbf{(b, e)} Top-$K$ gating-score entropy, where a larger value means a more even mixture over selected experts. \textbf{(c, f)} Full-pool selection-score entropy, where a larger value means a flatter selection-score distribution.}
    \label{fig:infer-util}
\end{figure*}

\paragraph{Expert utilization at inference.}
We evaluate expert utilization in continued-pretrained models by recording layer-wise loads and router scores in prefill mode on nine downstream benchmarks. Each layer value is averaged over the benchmarks. The evaluated models use Top-$8$-of-$256$ and Top-$10$-of-$768$ routing. Models within each setting share the same architecture. ID Balancing uses $K_i=K_d=6\times10^{-6}$ during continued pretraining. Fig.~\ref{fig:infer-util}(a,\,d) reports the \emph{inactive-expert ratio}, the fraction of experts receiving no evaluation token. Under Top-$8$-of-$256$ routing, its average ratio is $3.1\%$, compared with $3.5\%$ for DeepSeek's loss-free method, $4.1\%$ for Quantile Balancing, and $4.7\%$ for Auxiliary loss. Under Top-$10$-of-$768$ routing, it reduces the average ratio from $8.4\%$ to $6.4\%$ and the first-layer ratio from $9.3\%$ to $2.6\%$ relative to Auxiliary loss. These results show fewer unused experts on the evaluation data, consistent with the lower training-time MinVio in Fig.~\ref{fig:perlayer-vio}. Under Top-$8$-of-$256$ routing, inactive ratios generally increase in deeper layers across methods. Fig.~\ref{fig:infer-util}(b,\,e) reports the top-$K$ gating-score entropy, $H_{\text{top-}K}=-\sum_{i=1}^{K}w_i\log w_i$ in nats, where $w_i$ is the normalized combination weight of selected expert $i$. Higher entropy indicates a more even mixture over selected experts. Under Top-$8$-of-$256$ routing, the loss-free methods have slightly higher entropy than Auxiliary loss on average across layers, while differences among DeepSeek's loss-free method, Quantile Balancing, and ID Balancing are small. Fig.~\ref{fig:infer-util}(c,\,f) reports the full-pool selection-score entropy, $H_{\text{full}}=-\sum_{e=1}^{E}p_e\log p_e$. For Auxiliary loss, $p_e$ is the softmax probability over router logits. For the loss-free methods, it is obtained by clamping the biased sigmoid score $\sigma(z_e)+b_e$ to non-negative values and normalizing across experts. Higher entropy indicates a flatter score distribution. Its absolute values are not directly comparable across these score families. Under Top-$8$-of-$256$ routing, the loss-free score distributions become sharper with depth, while Auxiliary loss shows the opposite trend.

\begin{table*}[t]
\centering
\caption{\textbf{An integral gain of $K_i=6\times10^{-3}$ yields the lowest worst-case backbone MaxVio among the tested gains.} Results use a $0.87$B-active/$18.9$B-total model with Top-$3$-of-$768$ routing over $30$k steps. We fix $K_d=0$ and vary $K_i\in\{3,6,9\}\times10^{-3}$. Increasing $K_i$ to $9\times10^{-3}$ further reduces training-average backbone violations but worsens MTP balance. Lower values are better for all metrics.}
\label{tab:exp-ablation}
\resizebox{\textwidth}{!}{%
\begin{tabular}{l*{11}{c}}
\toprule
 & \multicolumn{6}{c}{\textbf{Backbone}} & \multicolumn{5}{c}{\textbf{MTP module}} \\
\cmidrule(lr){2-7}\cmidrule(lr){8-12}
\textbf{Setting} & LM & \multicolumn{3}{c}{MaxVio $\downarrow$} & \multicolumn{2}{c}{MinVio $\downarrow$} & \multicolumn{3}{c}{MaxVio $\downarrow$} & \multicolumn{2}{c}{MinVio $\downarrow$} \\
\cmidrule(lr){3-5}\cmidrule(lr){6-7}\cmidrule(lr){8-10}\cmidrule(lr){11-12}
 & loss & Last1k & Avg. & Worst & Last1k & Avg. & Last1k & Avg. & Worst & Last1k & Avg. \\
\midrule
\multicolumn{12}{l}{\cellcolor{mgray}\textbf{Integral gain $K_i$, integral term only, $K_d{=}0$}} \\
  
  $K_i=3\times10^{-3}$ & \cellcolor{loss7}\textbf{1.7422} & \textbf{0.6759} & \cellcolor{mvio3}0.9459 & \cellcolor{mvio4}23.26 & \cellcolor{mvio3}\textbf{0.5190} & \cellcolor{mvio3}0.5777 & \textbf{0.6375} & \textbf{0.7991} & \textbf{9.14} & \textbf{0.5603} & \cellcolor{mvio3}0.5775 \\
  $K_i=6\times10^{-3}$ & 1.7431 & \cellcolor{mvio2}0.6924 & \cellcolor{mvio2}0.8003 & \textbf{14.96} & \cellcolor{mvio3}0.5233 & \cellcolor{mvio3}0.5527 & \cellcolor{mvio3}0.8557 & \cellcolor{mvio3}0.8862 & \cellcolor{mvio4}15.18 & \cellcolor{mvio3}0.5848 & \textbf{0.5714} \\
  $K_i=9\times10^{-3}$ & \cellcolor{loss4}1.7429 & \cellcolor{mvio2}0.6831 & \textbf{0.7872} & \cellcolor{mvio4}20.13 & \cellcolor{mvio3}0.5263 & \textbf{0.5436} & \cellcolor{mvio3}0.9222 & \cellcolor{mvio3}1.0388 & \cellcolor{mvio4}35.56 & \cellcolor{mvio3}0.6279 & \cellcolor{mvio3}0.5800 \\
\bottomrule
\end{tabular}%
}
\end{table*}

\begin{table*}[t]
\centering
\caption{\textbf{Larger derivative gains reduce backbone overload over $1$--$5$k steps but increase MTP overload.} Results use a $0.87$B-active/$18.9$B-total model with Top-$3$-of-$768$ routing over $30$k steps. We fix $K_i=6\times10^{-3}$ and vary $K_d\in\{0,3,6,12\}\times10^{-3}$, with $K_d=0$ as the integral-only baseline. Load metrics summarize the indicated early intervals and the full training window (Avg.).}
\label{tab:dterm-phase-ablation}
\resizebox{\textwidth}{!}{%
\begin{tabular}{l*{11}{c}}
\toprule
 & \multicolumn{7}{c}{\textbf{Backbone}} & \multicolumn{4}{c}{\textbf{MTP module}} \\
\cmidrule(lr){2-8}\cmidrule(lr){9-12}
\textbf{Setting} & LM & \multicolumn{2}{c}{$0$--$1$k steps} & \multicolumn{2}{c}{$1$--$5$k steps} & \multicolumn{2}{c}{Avg.} & \multicolumn{2}{c}{$0$--$5$k steps} & \multicolumn{2}{c}{Avg.} \\
\cmidrule(lr){3-4}\cmidrule(lr){5-6}\cmidrule(lr){7-8}\cmidrule(lr){9-10}\cmidrule(lr){11-12}
 & loss & MaxVio $\downarrow$ & MinVio $\downarrow$ & MaxVio $\downarrow$ & MinVio $\downarrow$ & MaxVio $\downarrow$ & MinVio $\downarrow$ & MaxVio $\downarrow$ & MinVio $\downarrow$ & MaxVio $\downarrow$ & MinVio $\downarrow$ \\
\midrule
\multicolumn{12}{l}{\cellcolor{mgray}\textbf{Derivative gain $K_d$, fixed $K_i=6\times10^{-3}$}} \\
  $K_d=0$ (I only) & \cellcolor{loss1}1.7431 & \cellcolor{mvio4}2.0020 & \cellcolor{mvio4}0.7955 & \cellcolor{mvio4}0.9059 & \cellcolor{mvio3}0.5989 & \cellcolor{mvio4}0.8003 & \cellcolor{mvio4}0.5527 & \cellcolor{mvio1}\textbf{1.4965} & \cellcolor{mvio3}0.6222 & \cellcolor{mvio1}\textbf{0.8862} & \cellcolor{mvio2}0.5714 \\
  $K_d=3\times10^{-3}$ & \cellcolor{loss3}1.7428 & \cellcolor{mvio2}1.8900 & \cellcolor{mvio3}0.7853 & \cellcolor{mvio3}0.8869 & \cellcolor{mvio4}0.5992 & \cellcolor{mvio2}0.7910 & \cellcolor{mvio2}0.5485 & \cellcolor{mvio2}1.7454 & \cellcolor{mvio2}0.6201 & \cellcolor{mvio2}0.9459 & \cellcolor{mvio4}0.5799 \\
  $K_d=6\times10^{-3}$ & \cellcolor{loss7}\textbf{1.7426} & \cellcolor{mvio3}1.9330 & \cellcolor{mvio2}0.7791 & \cellcolor{mvio2}0.8612 & \cellcolor{mvio2}0.5899 & \cellcolor{mvio3}0.7931 & \cellcolor{mvio1}\textbf{0.5480} & \cellcolor{mvio3}1.7724 & \cellcolor{mvio4}0.6336 & \cellcolor{mvio3}0.9698 & \cellcolor{mvio3}0.5782 \\
  $K_d=12\times10^{-3}$ & \cellcolor{loss5}1.7427 & \cellcolor{mvio1}\textbf{1.8100} & \cellcolor{mvio1}\textbf{0.7644} & \cellcolor{mvio1}\textbf{0.8131} & \cellcolor{mvio1}\textbf{0.5833} & \cellcolor{mvio1}\textbf{0.7812} & \cellcolor{mvio3}0.5501 & \cellcolor{mvio4}2.0382 & \cellcolor{mvio1}\textbf{0.6048} & \cellcolor{mvio4}1.0922 & \cellcolor{mvio1}\textbf{0.5660} \\
\bottomrule
\end{tabular}%
}
\end{table*}

\subsection{Ablations}
\label{sec:exp:ablation}

\begin{figure*}[t]
    \centering
    \includegraphics[width=\linewidth]{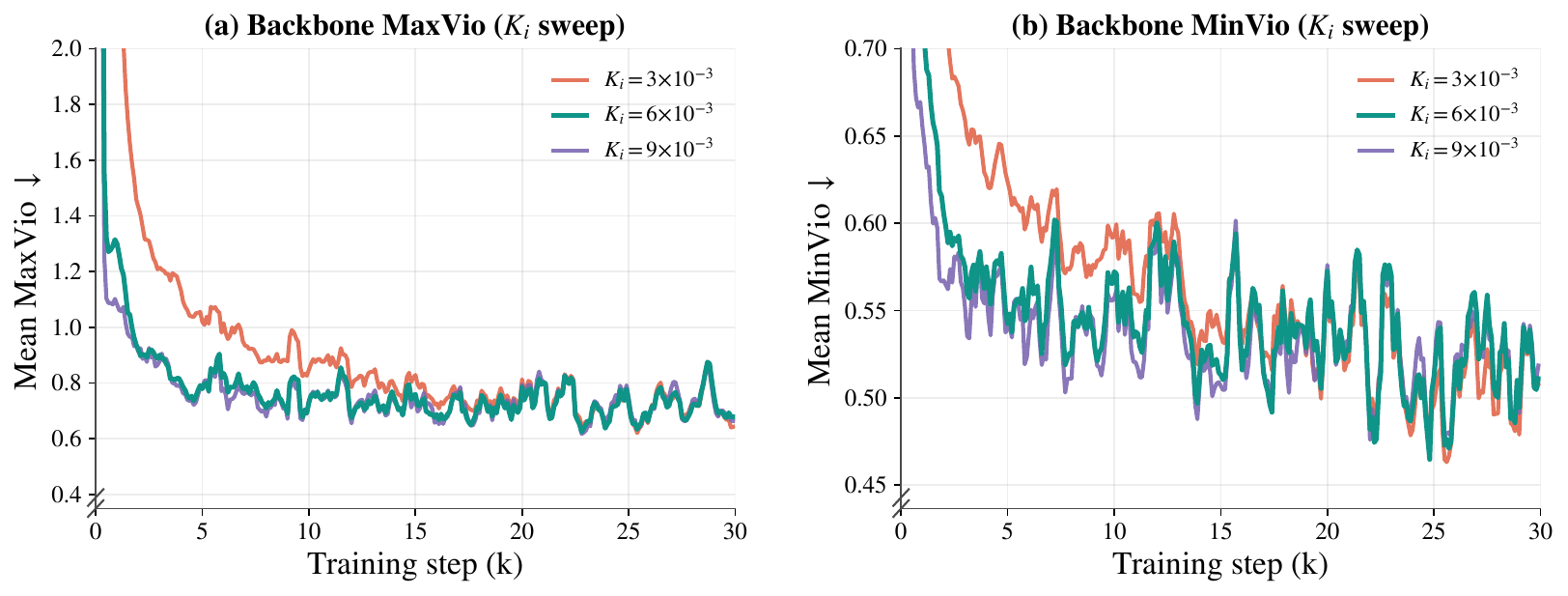}
    \caption{\textbf{A small integral gain responds slowly to the early imbalance, while larger gains converge to similar backbone trajectories.} Results use Top-$3$-of-$768$ routing over $30$k steps with $K_d=0$. We compare $K_i\in\{3,6,9\}\times10^{-3}$. \textbf{(a)} Mean backbone MaxVio ($\downarrow$). \textbf{(b)} Mean backbone MinVio ($\downarrow$).}
    \label{fig:ki-ablation}
\end{figure*}

\paragraph{Integral gain $K_i$ in isolation.}
With $K_d=0$, we vary $K_i\in\{3,6,9\}\times10^{-3}$. Fig.~\ref{fig:ki-ablation} shows that the smallest gain corrects early backbone imbalance more slowly, while the trajectories become similar later. Increasing $K_i$ from $3\times10^{-3}$ to $6\times10^{-3}$ reduces training-average backbone MaxVio from $0.9459$ to $0.8003$ and worst-case MaxVio from $23.26$ to $14.96$ (Tab.~\ref{tab:exp-ablation}). A further increase to $9\times10^{-3}$ lowers the average only slightly, to $0.7872$, but raises worst-case backbone MaxVio to $20.13$ and worst-case MTP MaxVio from $15.18$ to $35.56$. The smallest gain retains the lowest LM loss and several final-stage and MTP load metrics. We therefore choose $K_i=6\times10^{-3}$ for its stronger early correction and lowest worst-case backbone MaxVio, while avoiding the larger MTP violations at $9\times10^{-3}$.

\begin{figure*}[t]
    \centering
    \includegraphics[width=\linewidth]{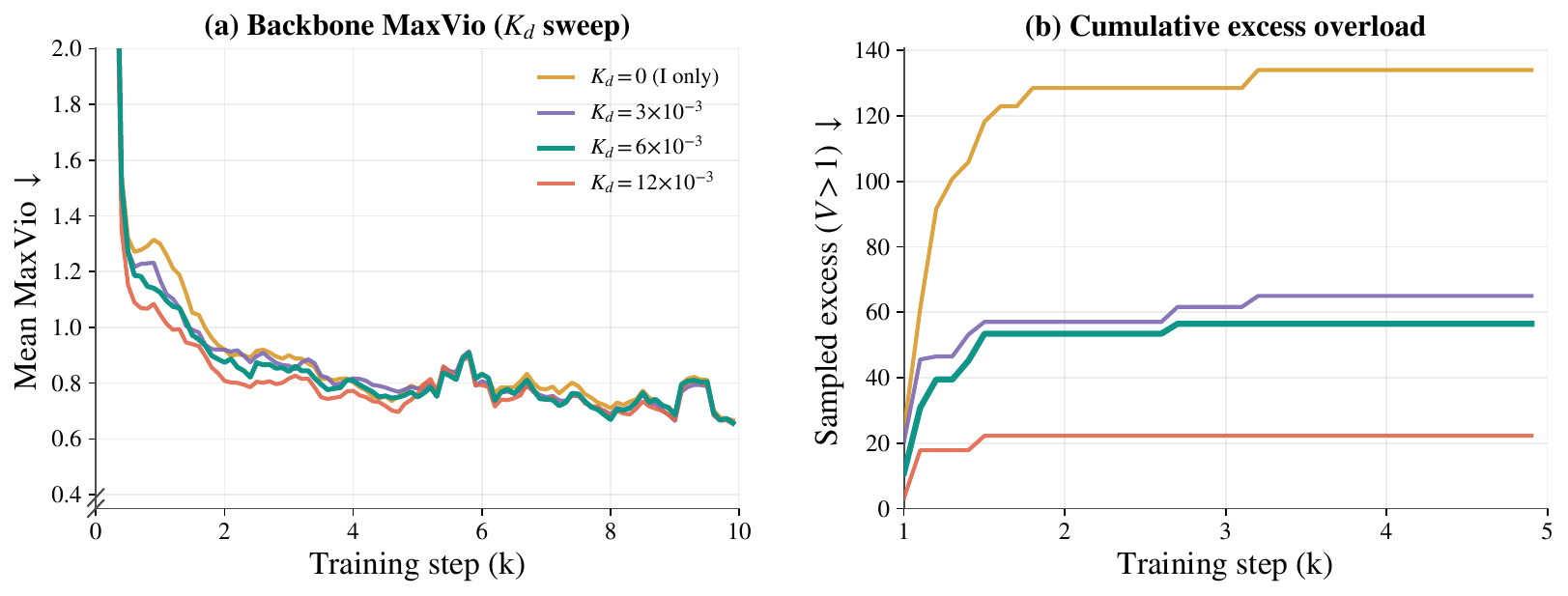}
    \caption{\textbf{The derivative term reduces early backbone overload.} Results use Top-$3$-of-$768$ routing with $K_i=6\times10^{-3}$ and $K_d\in\{0,3,6,12\}\times10^{-3}$, where $K_d=0$ is the integral-only baseline. Let $V_t$ denote mean MaxVio over the $20$ backbone layers. \textbf{(a)} $V_t$ over the first $10$k steps, smoothed with a five-point moving average. \textbf{(b)} Cumulative excess above $V_t=1$ over $1$--$5$k steps.}
    \label{fig:kd-ablation}
\end{figure*}

\paragraph{Derivative gain $K_d$ and early load control.}
With $K_i=6\times10^{-3}$, we vary $K_d\in\{0,3,6,12\}\times10^{-3}$, using $K_d=0$ as the integral-only baseline. Adding the derivative term improves early backbone load control, while later trajectories remain similar (Fig.~\ref{fig:kd-ablation}(a)). The default $K_d=6\times10^{-3}$ reduces mean backbone MaxVio from $2.0020$ to $1.9330$ over $0$--$1$k steps and from $0.9059$ to $0.8612$ over $1$--$5$k steps, with lower MinVio in both intervals (Tab.~\ref{tab:dterm-phase-ablation}). Over $1$--$5$k steps, it also reduces the sampled cumulative excess above layer-mean MaxVio $V_t=1$ by $57.9\%$ relative to the integral-only baseline, using unsmoothed observations (Fig.~\ref{fig:kd-ablation}(b)). These early gains come with a trade-off in MTP balance. Larger derivative gains reduce backbone overload over $1$--$5$k steps but increase MTP overload. At the default gain, training-average MTP MaxVio rises from $0.8862$ to $0.9698$, while training-average backbone MaxVio improves only modestly, from $0.8003$ to $0.7931$. We choose $K_d=6\times10^{-3}$ because it provides stronger backbone control over $1$--$5$k steps than $3\times10^{-3}$, with less MTP overload than $12\times10^{-3}$. It also yields the lowest LM loss and training-average backbone MinVio among the tested gains.

\section{Related Work}

\paragraph{Sparse mixture-of-experts.}
Transformer architectures and large-scale pretraining have driven progress in language modeling~\citep{vaswani2017attention,radford2019language,brown2020language}, followed by instruction tuning and conversational models~\citep{ouyang2022training,chatgpt,achiam2023gpt}. Sparse MoE models extend this scaling direction by increasing total capacity through conditional expert activation, with a smaller increase in per-token computation~\citep{du2022glam,rajbhandari2022deepspeed,muennighoff2025olmoe,puigcerver2024sparse,he2024mixture,yang2025qwen3}. Switch Transformer~\citep{fedus2022switch} and GShard~\citep{lepikhin2020gshard} use Top-$1$ and Top-$2$ routing, respectively, while BASE Layers~\citep{lewis2021base}, Hash Layers~\citep{roller2021hash}, and Expert Choice~\citep{zhou2022mixture} explore alternative token--expert assignments. Sparse activation also extends beyond standard feed-forward experts: MoH~\citep{jin2024moh} applies selective activation to attention heads, and MoE++~\citep{jin2025moe++} combines feed-forward and zero-computation experts. Our work addresses the load-control challenge that arises as the active fraction of a Top-$K$ MoE decreases.

\paragraph{Balancing through an auxiliary loss.}
Auxiliary-loss methods encourage balanced routing by combining each expert's token fraction with its average router probability~\citep{fedus2022switch}. The resulting gradients update the same router weights as the language-modeling objective, introducing a trade-off between the two objectives~\citep{yu2020gradient,sener2018multi,liu2021conflict}. System-level approaches such as LocMoE~\citep{li2024locmoe} and MegaBlocks~\citep{gale2023megablocks} further address efficient MoE execution. ID Balancing controls expert selection through a separate bias update, without adding a balancing gradient to the training objective.

\paragraph{Auxiliary-loss-free balancing.}
Auxiliary-loss-free methods add an expert-specific bias for Top-$K$ selection and exclude it from the combination weights~\citep{wang2024auxiliary,liu2024deepseek}. DeepSeek's loss-free method updates this bias using the sign of the load error. Its fixed step discards error magnitude, creating a trade-off between correcting large errors and avoiding overshoot near balance. Kimi K3's Quantile Balancing~\citep{team2026kimi} instead computes a batch-dependent target bias from quantiles of router-score margins. At inference, the learned bias is frozen and routing uses ordinary Top-$K$ selection. We propose a generalized PID view that organizes these updates by how they use feedback over time: DeepSeek's loss-free method acts as fixed-step integral control, while Quantile Balancing acts as generalized proportional control. This view motivates ID Balancing, which combines magnitude-aware integral control with a worsening-gated derivative correction.

\section{Conclusion}

Effective load control is essential to using the capacity of larger, sparser MoE models. We propose a generalized PID view of auxiliary-loss-free balancing and use it to develop ID Balancing, which combines magnitude-aware integral control with a worsening-gated derivative correction and zero-mean bias centering. The method requires only $\mathcal{O}(E)$ token-count feedback per layer and adds no balancing gradient to the language-modeling objective. Experiments across routing sparsities, model sizes, and learning-rate settings show effective load control while maintaining competitive language-modeling and downstream performance. Reduced gains retain active load correction during continued pretraining, and further analyses show consistent backbone load control across layers and fewer inactive experts on evaluation data. These results support ID Balancing as a practical approach to load control for larger, sparser MoE models.

\section*{Limitations}
Our evaluation covers several routing sparsities, model sizes, and training conditions within a family of decoder-only MoE models using fixed Top-$K$ routing. Extending the evaluation to other architectures and routing methods is a natural next step. The accumulated worsening-gated derivative update is supported empirically. The scaling experiments compare complete model configurations that differ in depth, width, active and total parameters, and training duration, so they assess load control across configurations without isolating the effect of model size alone.

\bibliography{custom}
\bibliographystyle{colm2024_conference}

\appendix
\renewcommand{\thefootnote}{\fnsymbol{footnote}}
\renewcommand{\thetable}{\Alph{table}}
\renewcommand{\theequation}{\Alph{equation}}
\renewcommand{\thefigure}{\Alph{figure}}

\setcounter{table}{0}
\setcounter{section}{0}
\setcounter{figure}{0}
\setcounter{equation}{0}

\section{Relation to LongCat-Flash's Bias Update}
\label{app:longcat-integral}

LongCat-Flash~\citep{longcat2025technical} uses a PID-inspired expert-bias update to regulate the average computational budget of an MoE layer with zero-computation experts~\citep{jin2025moe++}. To compare its update with our integral term, let $E$ denote the number of FFN experts, $T$ the number of tokens in the global batch used for each bias update, $K$ the total number of experts selected per token, and $K_e$ the expected number of selected FFN experts. Using our notation for expert loads, its reported update for an FFN expert is:
\begin{equation}
    b_i^{(t+1)}=b_i^{(t)}
    +\mu\left(\frac{K_e}{KE}-\frac{n_i^{(t)}}{KT}\right),
    \label{eq:longcat-bias}
\end{equation}
where $\mu>0$ is the bias adaptation rate. The biases of zero-computation experts remain fixed.

Now consider the setting without zero-computation experts, where each token selects exactly $K$ FFN experts and no routing assignments are dropped. In this case, $K_e=K$ and $\bar n^{(t)}=KT/E$. Substituting these identities and the normalized error from Eq.~\ref{eq:pid-error} into Eq.~\ref{eq:longcat-bias} gives:
\begin{equation}
    \begin{aligned}
    b_i^{(t+1)}-b_i^{(t)}
    &=\mu\left(\frac{1}{E}-\frac{n_i^{(t)}}{KT}\right)\\
    &=\frac{\mu}{E}\left(1-\frac{n_i^{(t)}}{\bar n^{(t)}}\right)
    =\frac{\mu}{E}\,e_i^{(t)}.
    \end{aligned}
    \label{eq:longcat-integral-equivalence}
\end{equation}
Thus, under these assumptions, LongCat-Flash's update has the same form as the integral update in Eq.~\ref{eq:rapid-integral}, with $K_i=\mu/E$. Both accumulate load errors in the bias and scale each correction with the error magnitude. Their gain parameters differ by the expert-count factor $E$; any schedule for $\mu$ corresponds to the same schedule for $K_i$ divided by $E$.

This equivalence concerns the bias-update rule under matching routing rules, load statistics, and initialization. LongCat-Flash additionally uses a device-level load-balancing loss, while the complete ID Balancing update includes the worsening-gated derivative correction and bias re-centering in Eq.~\ref{eq:rapid-update}.

\section{Experimental Setup}

\subsubsection{Scaling Setups}
\label{app:scaling-setups}

\textbf{Model architectures.}
We evaluate three hybrid MoE configurations that cover different expert-pool and model scales. Their specifications are summarized in Tab.~\ref{tab:scaling-setups}. All three models combine full attention and gated linear attention~\citep{yang2023gated,yang2025gated,dao2024transformers,sun2023retentive}, with one full-attention layer every four Transformer layers. Each MoE layer contains one shared expert in addition to the routed experts. The input embedding and output head are untied.

\begin{table*}[t]
\centering
\caption{\textbf{Architectural specifications for the evaluated MoE models.} Parameter counts include the input embedding and untied output head, and exclude the MTP module. Active parameters include all dense and shared components together with the routed experts selected by Top-$K$.}
\label{tab:scaling-setups}
\resizebox{\textwidth}{!}{%
\begin{tabular}{lccc}
\toprule
\textbf{Specification} & \textbf{Small MoE} & \textbf{Medium MoE} & \textbf{Large MoE} \\
\midrule
Transformer layers & 20 & 28 & 28 \\
Hidden size & 1024 & 2048 & 2048 \\
Vocabulary size & 248320 & 248320 & 248320 \\
Softmax-attention heads & 16 & 16 & 16 \\
Query groups & 2 & 2 & 2 \\
Softmax-attention head dimension & 256 & 256 & 256 \\
Full-attention interval & 4 & 4 & 4 \\
Linear-attention key/value dimension & 128 / 128 & 128 / 128 & 128 / 128 \\
Linear-attention key/value heads & 8 / 32 & 16 / 32 & 16 / 32 \\
Per-expert FFN size & 384 & 512 & 512 \\
Routed experts & 768 & 256 & 768 \\
Shared experts & 1 & 1 & 1 \\
Top-$K$ & 3 / 5 / 10 & 8 & 10 \\
Tied input/output embeddings & No & No & No \\
Total parameters & 18.9B & 24.8B & 69.9B \\
Active parameters & 0.87B / 0.91B / 1.03B & 3.0B & 3.2B \\
\bottomrule
\end{tabular}%
}
\end{table*}

\end{document}